\documentclass[a4paper]{article}
\usepackage{etoolbox}

\usepackage{amsmath,amsthm}
\usepackage{mathtools}
\usepackage[dvipsnames,table]{xcolor}

\usepackage{natbib}
\usepackage{authblk}

\RequirePackage{crop}
\RequirePackage{graphicx}
\usepackage[labelfont=it]{caption}
\RequirePackage{amsmath}
\RequirePackage{array}
\RequirePackage{color}
\RequirePackage{xcolor}
\RequirePackage{amssymb}
\RequirePackage{flushend}
\RequirePackage{stfloats}
\RequirePackage[figuresright]{rotating}
\RequirePackage{chngpage}
\RequirePackage{totcount}
\RequirePackage{fix-cm}
\RequirePackage{hyperref}
\usepackage{doi}

\definecolor{linkcolor}{HTML}{991408}  % red
\definecolor{citecolor}{HTML}{2E7E2A}  % green
\definecolor{filecolor}{HTML}{131877}  % dark blue
\definecolor{menucolor}{HTML}{727500}  % yellow
\definecolor{runcolor} {HTML}{137776}  % teal
\definecolor{urlcolor} {HTML}{0a2bbf}  % blue
\hypersetup{colorlinks=true,linkcolor=linkcolor,citecolor=citecolor,filecolor=filecolor,menucolor=menucolor,runcolor=runcolor,urlcolor=urlcolor}
\newtoggle{arxiv}
\toggletrue{arxiv}% To build paper version, please use main.tex instead
\iftoggle{arxiv}{
  \usepackage[parfill]{parskip}

  \usepackage[%
    a4paper,
    inner=25mm,
    outer=25mm,% = marginparsep + marginparwidth + 5mm (between marginpar and page border)
    top=25mm,
    bottom=25mm,
    marginparsep=5mm,
    marginparwidth=40mm,
  ]{geometry}

  \newlength{\defbaselineskip}
  \setlength{\defbaselineskip}{\baselineskip}
  \setlength{\parskip}{6pt}%
  \setlength{\parindent}{0pt}%

  \RequirePackage[T1]{fontenc}
  \RequirePackage[tt=false, type1=true]{libertine}
  \RequirePackage[varqu]{zi4}
  \RequirePackage[libertine]{newtxmath}
}{}

\usepackage{booktabs}
\usepackage{makecell}
\usepackage{placeins}%FloatBarrier

\usepackage{multirow}

\usepackage{tikz}
\usepackage{float}
\usepackage{subcaption}
\usepackage{amssymb}% http://ctan.org/pkg/amssymb
\usepackage{pifont}% http://ctan.org/pkg/pifont
\newcommand{\xmark}{\ding{55}}%

\graphicspath{{Figures/}}

\usepackage{comment}

\newcommand{\mypara}[1]{\textbf{#1.}}
\newcommand{\first}[1]{\textbf{#1}}
\newcommand{\runnerup}[1]{\underline{#1}}
\newcommand{\selected}[1]{$\rightarrow$~\textit{#1}~$\leftarrow$}
\newcommand{\na}[1]{{\color{darkgray}{N/A}}}

\newcommand{\hreffoot}[2]{\href{#1}{#2}\footnote{\url{#1}}}
\newcommand{\appref}[1]{\hyperref[#1]{Appendix~\ref*{#1}}}

\definecolor{ForestGreen}{RGB}{34,139,34}

\title{BarcodeBERT: Transformers for Biodiversity Analyses}

\usepackage{fancyhdr}
\usepackage{multicol}
\makeatletter
\renewcommand\@author{
    \AB@authlist\\[\affilsep]
    \begin{minipage}{0.9\textwidth}
    \begin{multicols}{2}
        \raggedright
        \AB@affillist
    \end{multicols}
    \end{minipage}
    }
\makeatother

\author[1,$\ast$]{Pablo~Millan~Arias}
\author[1,$\ast$]{Niousha~Sadjadi}
\author[1,$\ast$]{Monireh~Safari}
\author[3,$\dagger$]{\\ZeMing~Gong}
\author[3,$\dagger$]{Austin~T.~Wang}
\author[6]{Joakim~Bruslund~Haurum}
\author[2,4]{Iuliia~Zarubiieva}
\author[2]{Dirk~Steinke}
\author[1,$\sharp$]{Lila~Kari}
\author[3,5]{Angel~X.~Chang}
\author[4,$\ddagger$]{Scott~C.~Lowe}
\author[2,4,$\ddagger$,$\sharp$]{Graham~W.~Taylor}

\affil[1]{University of Waterloo}
\affil[2]{University of Guelph}
\affil[3]{Simon Fraser University}
\affil[4]{Vector Institute}
\affil[5]{Alberta Machine Intelligence Institute (Amii)}
\affil[6]{Aalborg University and Pioneer Centre for AI}

\affil[$\ast$]{Joint first author}{ }
\affil[$\dagger$]{Joint second author}{ } 
\affil[$\ddagger$]{Joint senior author}{ }

\affil[$\sharp$]{Corresponding authors: \href{mailto:gwtaylor@uguelph.ca}{gwtaylor@uguelph.ca}, \phantom{$\sharp$}\href{mailto:lila@uwaterloo.ca}{lila@uwaterloo.ca}}

\date{}

\begin{document}

\maketitle
\begin{abstract}
In the global challenge of understanding and characterizing biodiversity, short species-specific genomic sequences known as DNA barcodes play a critical role, enabling fine-grained comparisons among organisms within the same kingdom of life. Although machine learning algorithms specifically designed for the analysis of DNA barcodes are becoming more popular, most existing methodologies rely on generic supervised training algorithms. We introduce BarcodeBERT, a family of models tailored to biodiversity analysis and trained exclusively on data from a reference library of 1.5\,M invertebrate DNA barcodes.  We compared the performance of BarcodeBERT on taxonomic identification tasks against a spectrum of machine learning approaches, including supervised training of classical neural architectures and fine-tuning of general DNA foundation models. Our self-supervised pretraining strategies on domain-specific data outperform fine-tuned foundation models, especially in identification tasks involving lower taxa such as genera and species. We also compared BarcodeBERT with BLAST, one of the most widely used bioinformatics tools for sequence searching, and found that our method matched BLAST's performance in species-level classification while being 55 times faster. 
Our analysis of masking and tokenization strategies also provides practical guidance for building customized DNA language models, emphasizing the importance of aligning model training strategies with dataset characteristics and domain knowledge. 
%To test our methodology, we select the taxonomic class of arthropods, a highly diverse and underexplored group. %The results show that our self-supervised pretraining strategies on domain-specific data outperform fine-tuned foundation models, especially in identification tasks involving lower taxa such as genera and species.
%This work provides practical guidance for researchers aiming to develop models tailored to specific biodiversity challenges.
The code repository is available at {\url{https://github.com/bioscan-ml/BarcodeBERT}}.

\end{abstract}
%\keywords{Biodiversity informatics, taxonomic classification, DNA barcode, machine learning, transformers, DNA language models}

\section{Introduction}
\label{sec:intro}
The task of estimating and understanding biodiversity on our planet remains a monumental challenge, as traditional methods of taxonomic analysis often struggle to keep pace with the rate of discovery and identification of new species. In this context, the search for highly expressive, short standardized genomic regions containing meaningful taxonomic information ({DNA barcodes}) has become prominent in biodiversity research over the past two decades \citep{HeberBarcodes2002, Miller2007, Hebert2016, Srivathsan2021}. Specifically, a 658-base-pair-long fragment of the Cytochrome c Oxidase Subunit I (COI) gene \citep{LuntBarcodesFirst1996} has emerged as the de facto DNA barcode for kingdom \textit{Animalia} \citep{DopheideDiversity2019} and has proven effective in addressing inherent taxonomic challenges. Particularly, barcodes can be used for fast and accurate queries to categorize novel specimens into existing taxa. Furthermore, in the absence of clear species boundaries, they can be used to systematically separate specimens into groups of closely related organisms. These clusters, known as operational taxonomic units (OTUs), correspond to groups of similar specimens and can be labelled using e.g.~a Barcode Index Number (BIN) \citep{RatnasinghamBOLD2007}. As it is defined systematically, such a BIN system overcomes ambiguities in traditional species labelling and thus accelerates biodiversity research.
Among the numerous taxonomic groups to which DNA barcoding is applicable, invertebrates, particularly arthropods, stand out as an incredibly diverse and taxonomically complex group \citep{Basset2012}, making them the focus of many methodological studies \citep{BadirliNEURIPS2021, BadirliJournal2023,gong2024bioscan}.  %with new species being discovered every year.
The diversity and taxonomic richness of this group require specialized algorithmic approaches that can capture the taxonomic structure of the data. Consequently, biodiversity researchers are increasingly turning to machine learning methods, including convolutional neural networks (CNNs) \citep{BadirliNEURIPS2021} and transformer models \citep{gong2024bioscan}, to scale taxonomic classification of arthropods and accelerate species discovery.

Transformer-based models, pretrained at scale with self-supervised learning (SSL), also referred to as ``foundation models,'' have found applications across diverse domains thanks to their effectiveness in learning from large unlabelled datasets \citep{GPT_3_NEURIPS2020, DeiTtouvron21a}. Such models are often task-agnostic and can perform well on a variety of downstream tasks after fine-tuning. Despite their success in other domains, their application for taxonomic identification using DNA barcodes has not yet been extensively explored. Moreover, most DNA-based foundation models primarily target human chromosomal DNA sequences \citep{zhou2023dnabert2, NucleotideTransformerDalla-Torre2023, Zhou2021DNABERT}, making them suboptimal for barcode data due to domain-shift between these data types. Though short, barcodes encode rich taxonomic information across over 265,000 animal species in the Barcode of Life Data System (BOLD).

We here aim to unlock the potential of transformer-based architectures for taxonomic identification of arthropod barcodes, providing insights that extend beyond broad, foundation-style approaches. We address the previously mentioned issues (i.e. the taxonomic complexity of arthropods, and the lack of specialized transformer models trained on DNA barcodes) by adopting a semi-supervised learning approach, followed by fine-tuning on high-quality labelled barcode data, demonstrating the value of targeted model development for specialized applications. 
We propose BarcodeBERT, a self-supervised method that leverages a reference library of 1.5M invertebrate barcodes \citep{1.5MdeWaard2019} and a masked language model (MLM) training strategy to effectively compute meaningful embeddings of the data, facilitating successful species-level classification of insect DNA barcodes in general scenarios. In addition to the classification of known species, our pretrained models can be used to generate embeddings for sequences from unseen taxa, enabling non-parametric classification at higher levels of the taxonomic hierarchy.

To summarize our contributions, we first investigate the impact of pretraining using a large and diverse DNA barcode dataset (1 million sequences, from more than 17,000 species, across  6,700 genera)   on generalization to other downstream tasks. Second, we compare BarcodeBERT against several baselines such as pretrained DNA foundation models (DNABERT \citep{Zhou2021DNABERT}, DNABERT-2 \citep{zhou2023dnabert2}, DNABERT-S\citep{zhou2024dnaberts}, the Nucleotide Transformer NT \citep{NucleotideTransformerDalla-Torre2023}, and HyenaDNA \citep{hyenadna}, a CNN baseline following the architecture introduced by \citep{BadirliNEURIPS2021}, and the widely used alignment-based method BLAST~\citep{altschul1990blast}. Third, our study provides actionable insights regarding tokenization strategies, optimal masking ratios, and the importance of application-specific pretraining for DNA language models.

Overall, BarcodeBERT outperforms all other foundation models in supervised species classification, matching BLAST's accuracy while being 55 times faster and more scalable. Moreover, a linear classifier trained on BarcodeBERT embeddings has $\sim$6\% higher species classification accuracy than the top-performing foundation model in this task. Lastly, the same embeddings can also be used for accurate genus classification using similarity searches, outperforming the top-performing foundation model by $\sim$30\%.

\vspace{-0.2cm}
\section{Related Work}\label{sec:Related_work}
The exponential growth of genomic datasets with the advent of high-throughput sequencing has both demanded and enabled a surge in classification tools for DNA sequences. Such tools are essential for large-scale biodiversity studies, where algorithmic approaches can expedite the taxonomic categorization of novel specimens. One intuitive approach is to embed sequences into a vector space where geometric distances approximate taxonomic similarities \citep{corso2021neural}. This allows for rapid comparisons between newly sequenced and labelled DNA, enabling accurate taxonomic assignments.

Many machine learning approaches, particularly in representation learning, have demonstrated considerable potential in biodiversity analyses as they can embed raw DNA data into an expressive lower-dimensional space.  Transformer-based models \citep{vaswani2017attention}  capture complex patterns within sequential data and have shown exceptional performance in various representation learning tasks across domains, either with or without supervision \citep{GPT_3_NEURIPS2020, DeiTtouvron21a, caron2021emerging}. These models are especially effective in learning from vast unlabelled datasets, making them ideal candidates for the analysis of genomic data, where obtaining high-quality annotations remains challenging. 

There has been a growing number of self-supervised learning-based DNA language models proposed recently, most of which are based on the transformer architecture and trained using the masked language model (MLM) objective. The first foundational model in this space, DNABERT, utilizes a BERT-based transformer architecture along with $k$-mer tokenization for genome sequence prediction tasks. Following DNABERT, other models have emerged, including the Nucleotide Transformer \citep{NucleotideTransformerDalla-Torre2023}, GENA-LM \citep{GENA_LM}, and HyenaDNA \citep{hyenadna}. While each model varies in architectural details, tokenization methods, and training data, their reliance on SSL and the MLM objective for pretraining remains a constant. HyenaDNA is a unique entry in this space as it uses a state-space model (SSM) based on the Hyena architecture \citep{poli2023hyena} and trains it for next-token prediction (a causal MLM).

The landscape of machine learning models specifically tailored for DNA barcodes is less developed. A recent study \citep{BadirliJournal2023} proposes a Bayesian framework based on CNNs which, when combined with visual information, achieves high accuracies in species-level identification of seen species and genus-level inference of novel species in a dataset of $\sim$32,000 insect DNA barcodes. This method uses supervised learning to compute meaningful embeddings that can be used as side information in a two-layer Bayesian zero-shot learning framework. Transformer methods have been introduced for the classification of fungal Internal Transcribed Spacer sequences without any self-supervision \citep{MycoAI_Duong}. 

Although there has been a growing number of SSL-based DNA language models proposed in the recent literature, our findings indicate that models pretrained on a diverse set of non-barcode DNA sequences underperform on downstream barcode tasks. This suggests that general DNA foundation models may struggle with the domain-specific characteristics of barcode data. In this study, we leverage barcode-specific training to improve both species-level classification accuracy and generalization to other taxonomic ranks. By grounding our approach in targeted data and architectural choices, we seek to advance the utility of machine learning in biodiversity research, moving beyond general off-the-shelf models trained to classify specimens into known taxa. Distinctly, our specialized models are not only capable of classifying known species but also can be used for taxonomic classification for species that are not present in the training set.
\vspace{-0.2cm}
\section{Methods}
\label{sec:methods}
In this section, we outline the key elements of our methodology. We begin with a detailed account of our datasets and data processing pipeline. We then describe the architectures and hyperparameters used in the development of BarcodeBERT.  

\subsection{Dataset}
We use a reference library for Canadian invertebrates \citep{1.5MdeWaard2019} for model training and testing, comprised of 1.5\,M DNA samples from BOLD \citep{RatnasinghamBOLD2007}. The dataset was further pre-processed and subdivided as described below.

\subsubsection{Data pre-processing}\label{s:data-preprocessing}  To ensure data integrity and consistency, we performed a series of pre-processing steps over this dataset. First, empty entries were removed. Then, following standard practices \citep{NucleotideTransformerDalla-Torre2023}, IUPAC ambiguity codes (non-\verb+ACGT+ symbols), including alignment gaps, were uniformly replaced with the symbol \verb+N+. Duplicate sequences were removed to avoid redundancy and increase the complexity of the training and pretraining tasks. Sequences with trailing \verb+N+'s were truncated. Finally, sequences falling below 200 base pairs or exhibiting over 50\% \verb+N+ characters were excluded. Our preprocessed dataset is available to download at \url{https://huggingface.co/datasets/bioscan-ml/CanadianInvertebrates-ML}.

\subsubsection{Data partitioning} 
\begin{figure}[!b]
\centering
\includegraphics[width=\iftoggle{arxiv}{0.85}{}\linewidth]{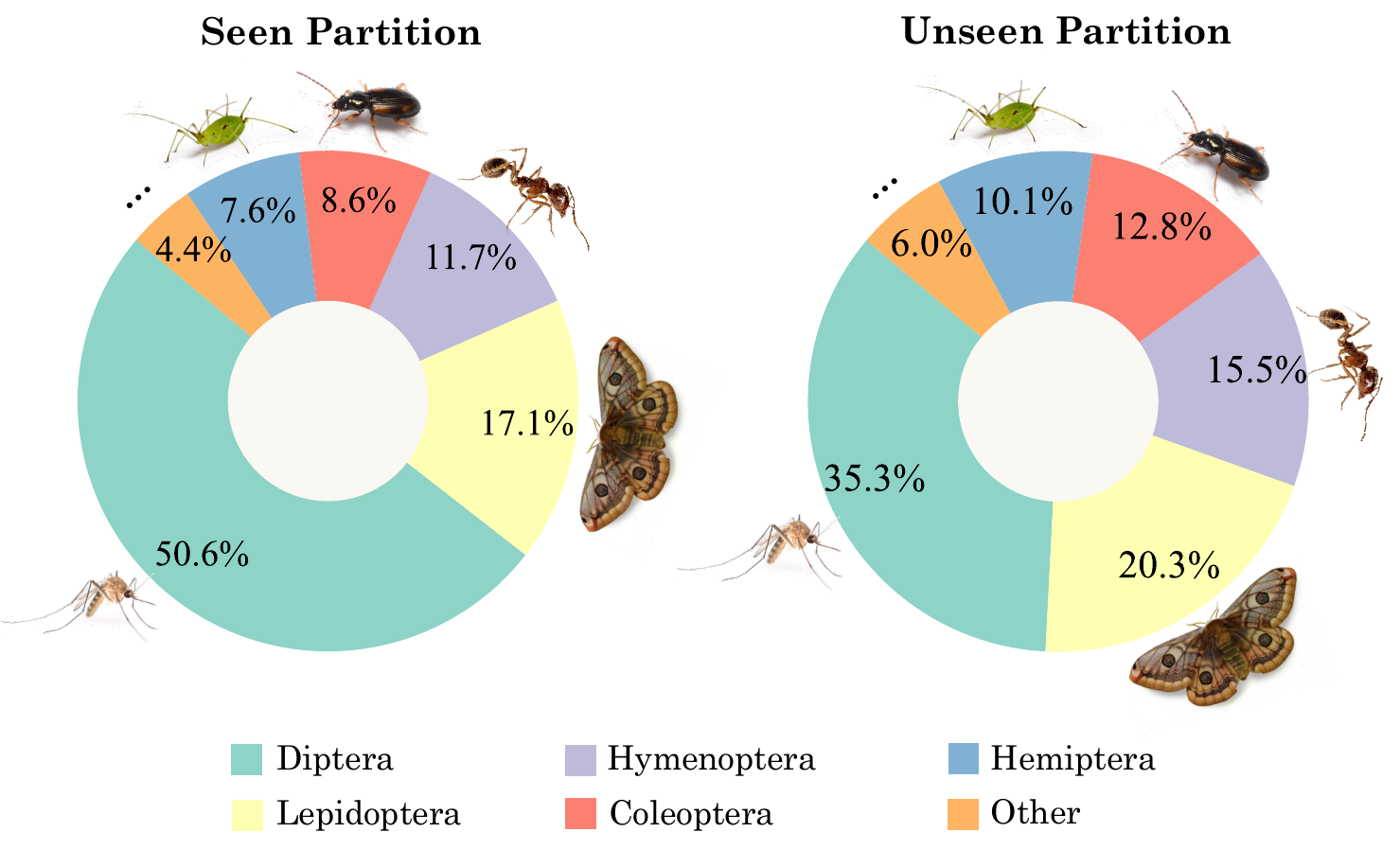}  

\caption{Distribution of orders in the \textit{Seen} (\textit{left}) and \textit{Unseen} (\textit{right}) partitions of the dataset. Icons: CC BY-SA, Wikimedia; Pro Content license, Canva.}
  \label{fig:order_distribution}
\end{figure}

After pre-processing, 965,289 barcode sequences from 17,464 invertebrate species, across 6,712 genera were obtained. The dataset was divided into three distinct partitions for different training and evaluation purposes: (\textit{i)~Seen}: This partition is intended for supervised learning pipelines, particularly to evaluate the model's ability to classify specimens from well-represented taxa. Comprised only of samples labelled to species-level, it includes 67,267 barcodes from 1,653 arthropod species, across 500 different genera, with each species represented by at least 20 and at most 50 barcodes. The partition is further split into training (70\%), testing (20\%), and validation (10\%) subsets. (\textit{ii)~Unseen}: This test partition was sampled to evaluate the models in real-world conditions where specimens from underrepresented species are frequently obtained. It only contains barcodes from ``rare'' species with fewer than 20 barcodes in the full reference dataset. Specifically, this partition contains 4,278 barcodes from 1,826 arthropod species, none of which are present in any other partition. Moreover, this partition contains all 500 genera labels present in the \textit{Seen} partition, with up to 20 barcodes sampled per genus. The label distribution shifts are shown in \autoref{fig:order_distribution}, with the \textit{Seen} partition reflecting the overall dataset’s distribution and the \textit{Unseen} partition exhibiting a greater diversity of rare genera.  (\textit{iii)~Pretrain}: This partition contains the remaining 893,744 barcode sequences from 14,794 invertebrate species across 6,679 genera. Note that only 35\% of the sequences in this partition contain full taxonomic annotations up to the species level. 
% The reader is referred to
See Supplement~A for more details on dataset composition.
\subsection{Proposed method: BarcodeBERT}
\begin{figure}[tbh]
\begin{minipage}{1.0\linewidth}
\centering
\includegraphics[width=\iftoggle{arxiv}{0.85}{0.90}\linewidth]{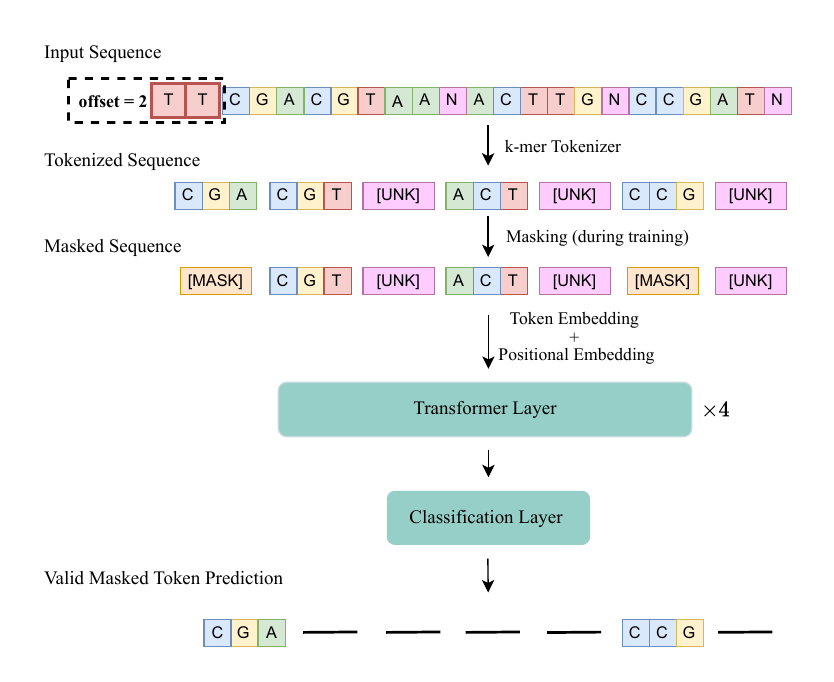}
\end{minipage}
\caption{ Architecture of BarcodeBERT, a transformer-based model employing a self-supervised learning strategy. The model is trained on non-overlapping $k$-mers from DNA barcode sequences. Any $k$-mer containing a character that is not in the nucleotide vocabulary is replaced by the \texttt{[UNK]} token. Pretraining involves masking out certain input parts using the \texttt{[MASK]} token and predicting these masked elements using a linear classification head. During training, the model selects a random offset ($0 \leq$ offset $< k$) for each sequence and begins tokenization from that position. This helps create more robust embeddings and increases resilience to potential mutations.}
\label{fig:arch}
\end{figure}

Inspired by Bidirectional Encoder Representations from Transformers (BERT)-like models, which convert sequence inputs into meaningful embedding vectors, BarcodeBERT is designed to encode DNA barcodes into informative embedding vectors for fast and effective comparisons.  This architecture's main building block is the transformer layer, with multi-head attention units playing a crucial role in capturing positional dependencies within each input sequence. Our model features four transformer layers, each with four attention heads, enabling a robust representation of the DNA barcode data while maintaining a manageable number of hyperparameters. \autoref{fig:arch} shows the details of BarcodeBERT architecture.

Before being fed as input to the model, each barcode is split into a sequence of tokens. After evaluating two of the most common tokenization strategies for DNA sequences, Byte Pair Encoding (BPE) \citep{zhou2023dnabert2, MycoAI_Duong} and $k$-mer tokenization \citep{Zhou2021DNABERT, NucleotideTransformerDalla-Torre2023}, we selected non-overlapping $k$-mer tokenization for BarcodeBERT (see the Ablation Studies section for more details). The token vocabulary includes all possible $k$-mer combinations derived from the nucleotide alphabet \{\verb+A+,\verb+C+,\verb+G+,\verb+T+\}, supplemented by two special tokens: \texttt{[MASK]} and \texttt{[UNK]}. The \texttt{[MASK]} token is utilized for masking \(k\)-mers during the pretraining phase, and \(k\)-mers containing any symbol that is not present in the nucleotide alphabet are assigned the \texttt{[UNK]} token. This results in a vocabulary size of $4^k + 2$.

A limitation of this tokenization strategy is its sensitivity to frame shifts. For example, the $k$-mer representation of the sequence \verb|GATCGA| differs entirely from that of \verb|CGATCGA|, even though the sequences differ by only a one-nucleotide shift. To address this issue and make our model robust to frame shifts that may occur in practice, we introduce a data augmentation step by randomly offsetting the sequence by a value ($0 \leq$ offset $< k$) during pretraining to improve generalization. Before tokenization, DNA barcodes are either padded or truncated at 660 nucleotides to ensure coverage of the barcode region in the COI gene \citep{elbrecht2019}. Finally, the tokenized sequences are fed to the model and encoded into a sequence of $768$-dimensional vectors.

Following self-supervised training, our model produces a whole barcode-level embedding vector by applying global average pooling over the sequence of $d$-dimensional output vectors, ignoring padding and any special tokens. During inference, the pipeline mirrors the training setup without the random offset: DNA barcodes are tokenized into non-overlapping $k$-mers and passed through the model, generating embeddings that capture meaningful taxonomic information across the entire sequence. BarcodeBERT is implemented using PyTorch and the Hugging Face Transformers library. During training, we focused exclusively on masked token prediction, masking 50\% of the input tokens and optimizing the network with a cross-entropy loss. We optimize the model parameters by gradient descent using the AdamW \citep{AdamW} optimizer with weight decay set to $1\times10^{-5}$ and a OneCycle schedule with maximum learning rate of $1\times10^{-4}$. Additionally, we performed experiments across different $k$-mer lengths (\(2 \leq k \leq 8\)) to observe the impact of $k$-mer length on embedding quality.
\section{Experiments}
\label{sec:results}
In this section, we present our evaluation framework and evaluate the performance of BarcodeBERT against the baseline models across several tasks. Additionally, we present a series of ablation studies to justify our design choices and analyze the impact of key hyperparameters on the model's performance.

\subsection{Experimental setup}
To explore the applicability of our model for DNA barcode-based biodiversity analyses,  we employ different SSL evaluation strategies \citep{balestriero2023cookbook} and contrast its performance against the baselines. First, we evaluate our models in a ``closed-world'' setting where the goal is to classify DNA sequences into known taxa.

\noindent \mypara{Fine-tuning} Pretrained models are fine-tuned on the training subset of the \textit{Seen} partition and evaluated on the test subset. This task assesses the ability of models to perform species-level classification with full access to labelled training data. 

\noindent \mypara{Linear probing} To evaluate the quality of pretrained embeddings, the backbone of the models is frozen, and a linear classifier is trained on the training subset of the \textit{Seen} partition. The final classifier is evaluated on the test subset, providing insights into the effectiveness of the embeddings without extensive task-specific training. 

\noindent \mypara{1-NN probing} This task evaluates model generalization to new species within known genera. Using cosine similarity, we perform 1-NN probing at the genus level with the training subset of the \textit{Seen} partition as the reference set and the \textit{Unseen} partition as the query set. 

Second, our models are evaluated
in an ``open-world'' setting where the goal is to group sequences, including those from unknown species, based on shared features.

\noindent \mypara{BIN reconstruction} We merge the test subset of the \textit{Seen} partition with the \textit{Unseen} partition and evaluate the model's ability to reconstruct Barcode Index Numbers (BINs) using embeddings generated without fine-tuning in a zero-shot clustering (ZSC) task \citep{zsc-Lowe-2024}. This task assesses how well the embeddings capture the hierarchical structure of taxonomic relationships, including rare or unclassified species.

%\subsubsection{Learning Embeddings for Zero-Shot Image Classification}
Finally, we evaluate the utility of learned DNA embeddings as auxiliary information in multi-modal learning.

\noindent \mypara{Bayesian zero-shot learning} We selected a species-level image classification task using the INSECT dataset \citep{BadirliNEURIPS2021}.
This is a small multimodal dataset designed for zero-shot classification of images from unseen species using DNA as auxiliary information. DNA embeddings generated by the models are paired with pre-extracted image features to classify species in a zero-shot setup. We evaluate both embeddings from pretrained and fine-tuned models on the species classification task from the INSECT dataset. Following \citep{BadirliNEURIPS2021}, the Bayesian zero-shot learning (BSZL) framework uses image features as priors and DNA embeddings as side information. For unseen species, the $K$-nearest seen species in the DNA embedding space are used to define local priors, allowing the Bayesian model to generate posterior predictive distributions for unseen categories. To ensure a fair comparison with prior work, image features are pre-extracted using ResNet-101 \citep{HeResNet2015}. Hyperparameter tuning for the Bayesian model is performed using the same grid search space as in \citep{BadirliNEURIPS2021}.  

\subsection{Results}
In this section, we describe our results on two evaluation tasks: DNA-specific tasks, designed to assess model performance in both open- and closed-world taxonomic settings; and zero-shot image classification using DNA embeddings.
\subsubsection{DNA-specific categorization tasks}
\begin{figure}[bth]
\centering
\vspace{-5mm}
\includegraphics[width=\iftoggle{arxiv}{0.75}{0.95}\linewidth]{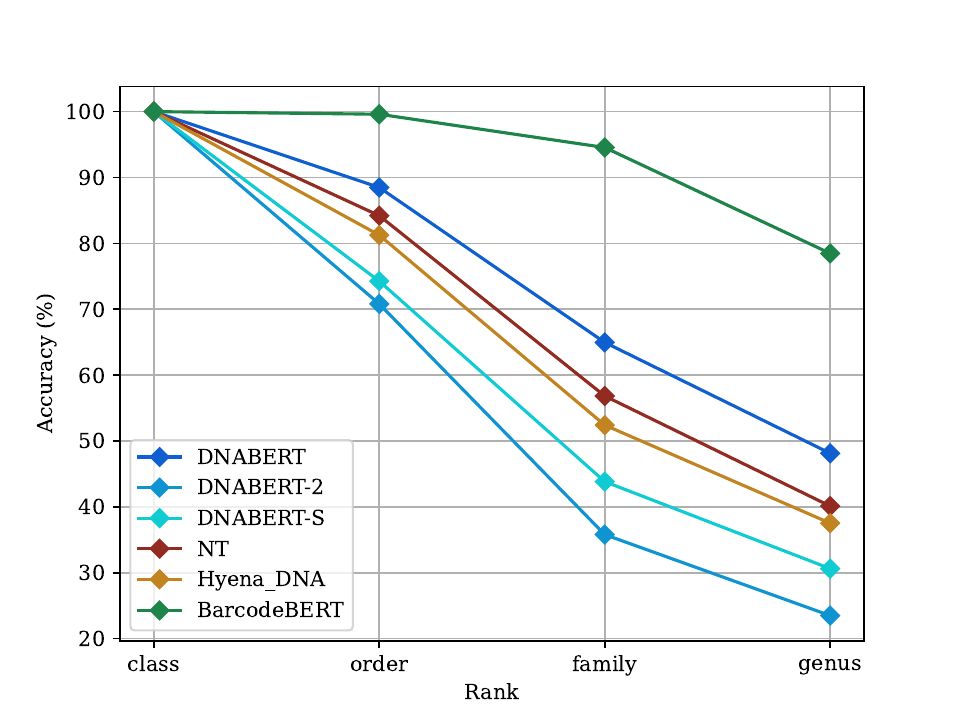}
\caption{Comparison of different DNA foundation models on the task of 1-NN probing at different taxonomic levels. The query set contains DNA barcodes from species not present in the key set, and none of the models have undergone fine-tuning.}
\label{fig:1NN-Taxa}
\end{figure}

Our evaluation (\autoref{tab:main_results_table}) compares several models across species-level and genus-level DNA-specific categorization tasks (fine-tuning, linear probing, 1-NN probing, BIN reconstruction through ZSC). For species-level classification, a BLAST search yielded 99.7\% accuracy based on the best hit. The performance of all fine-tuned deep learning-based models is comparable to this baseline, and all transformer models outperform the CNN model as well. \mbox{DNABERT-2}, DNABERT-S and BarcodeBERT all reached over 99.7\% accuracy. Notably, only BarcodeBERT continues to closely match BLAST's performance using a linear classifier, highlighting its strength in encoding meaningful features from raw data. 
In genus-level 1-NN probing, BarcodeBERT achieves the highest accuracy (78.5\%) among the deep learning-based models---more than double that of the same architecture without pretraining---demonstrating that the pretrained model produces richer embeddings, making BarcodeBERT far more robust for similarity searches of sequences from novel taxa. BLAST, however, performs best in this task (83.9\%). This indicates that without fine-tuning, BarcodeBERT captures coarser taxonomic distinctions but is limited in representing the full hierarchical taxonomic structure as shown in \autoref{fig:1NN-Taxa}. In addition to accuracy, we computed the F1-score for each model across these tasks, and observed similar trends (see Supplement D.8).

The ZSC task provides additional insights into the model's understanding of the hierarchical taxonomic structure. High performance in ZSC alone indicates a learned representation's ability to finely distinguish between closely related clusters (BINs) without necessarily capturing the higher-level taxonomy. In contrast, strong 1-NN performance at higher taxonomic levels but lower ZSC accuracy suggests that the model understands the overall topology of the hierarchical taxonomic structure, even if it lacks the granularity needed for precise clustering. DNABERT and BarcodeBERT exhibit this distinction, with BarcodeBERT achieving a more balanced performance across tasks, making it the more versatile model for comprehensive DNA barcode analysis.

\begin{table*}[!htp]\centering
\caption{Classification accuracy of DNA barcode models under different SSL evaluation strategies and different efficiency metrics. The baselines are divided into three groups: alignment-based techniques, BLAST; a deep learning-based non-SSL CNN baseline; and off-the-shelf DNA foundation models pretrained on non-barcode data. These are compared against BarcodeBERT, which is specifically pretrained on DNA barcode-based datasets. For BarcodeBERT we used the best configuration of $k = 4$, with 4 attention heads, and 4 layers. The DNABERT model supported variable stride length, and we show in parentheses the optimal $k$-mer value that yielded the best results. As an ablation, we also indicate the performance of the BarcodeBERT architecture when trained fully-supervised, without the self-supervised pretraining stage. Additionally, we show the throughput-per-second (TPS) of the encoders, and the total duration of the classification tasks. Numbers in boldface indicate the \first{best result} across each task, and {underlined} indicates \runnerup{second place.}}\label{tab:main_results_table}
\iftoggle{arxiv}{\small}{\scriptsize}
\resizebox{\ifdim\textwidth>\width\width\else\textwidth\fi}{!}{
\begin{tabular}{@{}lrrrrrrrr@{}}
\toprule
& & &\multicolumn{3}{c}{\makecell{Species-level acc (\%)\\ of seen species}} &\multicolumn{2}{c}{\makecell{Genus-level 1-NN probe\\ of unseen species}} & \makecell{BIN reconstruction \\ accuracy (\%)} \\\cmidrule(l){4-6} \cmidrule(l){7-8} \cmidrule(l){9-9}
Model &\#Param. &TPS (seq/s) &Finetuned &Linear probe &Dur (s) &Acc (\%) &Dur (s)&ZSC probe \\\midrule
BLAST &\na{} &\na{} &\multicolumn{2}{c}{\textit{\textbf{99.7}}$^{*}$} &1495 &\first{83.9} &602 &\na{} \\
\midrule
CNN encoder &1.8\,M &\runnerup{934} &98.2 & \na{} &\runnerup{13} &56.0 &\first{55}& \na{} \\
\midrule
DNABERT ($k\!=\!6$) &88.1\,M &50 &99.5 &\runnerup{98.1} & 248 &48.1 &1021&{\bf 81.5} \\
DNABERT-2 &118.9\,M &134 &\first{99.7} &95.7 & 101&23.5 & 381 &51.0 \\
DNABERT-S &117.1\,M &134 &\first{99.7} & 96.5 &101 &30.6& 381 &62.8 \\
HyenaDNA-tiny (d256) &1.6\,M &\first{1167} &99.4 &93.5 & \first{11} &50.6& \runnerup{76} &52.8 \\
Nucleotide Transformer &55.9\,M &95 &99.5 &96.2 & 140&40.1 & 536 &28.5 \\
\midrule
BarcodeBERT w/o pretraining & 29.1\,M & 484 & 99.5 & {\na{}} & 27 & 37.9 & 108 & \na{} \\ % Unfair ZSC = 81.2
BarcodeBERT (4-4-4) &29.1\,M &484 &\first{99.7} &\first{99.2} & 27&\runnerup{78.5} & 108 &\runnerup{79.9} \\
\bottomrule
\end{tabular}
}
\iftoggle{arxiv}{\vspace{1mm}}{\vspace{-1mm}}
\begin{flushleft}
{
\iftoggle{arxiv}{\footnotesize}{\scriptsizetiny}
$^{*}$BLAST is a deterministic algorithm without any learning component (see \iftoggle{arxiv}{\appref{a:methodology}}{Supplement~D.1} for details). Consequently,  species classification accuracy does not correspond to fine-tuning or linear probing, and it is only included in the table for reference.
}
\end{flushleft}
\iftoggle{arxiv}{}{\vspace{-2mm}}
\end{table*}

Two efficiency measurements are included: throughput, defined exclusively for deep-learning-based models as the number of sequences processed per second, and total runtime for classification pipelines to ensure a fair comparison with alignment-based baselines. In terms of throughput, HyenaDNA({\it tiny}) showcases the capabilities of state space models, demonstrating high throughput with fewer parameters. However, its classification performance is lower compared to BarcodeBERT and DNABERT-2. In total run time, our results indicate that subquadratic methods like the CNN baseline and HyenaDNA perform genus-level similarity searches {(1-NN~probe)} 13\texttimes{} faster than BLAST, while BarcodeBERT is 5.6 \texttimes{} faster than BLAST and outperforms other transformer models at this task. For species-level classification pipelines that do not include the computation of the training embeddings, transformer-based models demonstrate clear advantages over traditional baselines in terms of running time. Notably, BarcodeBERT, with a moderate parameter count, matches BLAST's high classification accuracy ($99.7\%$) with a 55\texttimes{} faster running time, thus providing a well-rounded option for large-scale DNA barcode applications.  All efficiency experiments were conducted using an Intel(R) Xeon(R) CPU @ 2.20GHz processor and a Quadro RTX 6000 GPU. Additional details on resource consumption can be found in Supplement~C.

\subsubsection{Zero-shot image classification using DNA embeddings}
We use the Bayesian zero-shot learning task to evaluate the quality of the DNA feature embeddings, by assessing their effectiveness when used as side information for classifying images to species on the INSECT dataset. We consider the embeddings directly from the pretrained models and also after fine-tuning. The accuracy for seen and unseen test species and the harmonic mean are presented in \autoref{bzsl-result-table}. Even without fine-tuning, BarcodeBERT substantially outperforms DNABERT and DNABERT-2 on unseen species, regardless of whether they had been fine-tuned previously. BarcodeBERT achieves similar performance to the reported baseline CNN results \citep{BadirliNEURIPS2021} and improves on the harmonic mean score by 1.2\% and unseen accuracy by 1.9\%, respectively. Our results demonstrate that in the zero-shot learning task of predicting insect species, employing BERT-like models that have also been trained on insect DNA barcodes as DNA encoders can improve performance over CNNs and general DNA foundation models.

\begin{table*}[t]
\centering
  \caption{Evaluation of DNA barcode models in a Bayesian zero-shot learning task on the INSECT dataset. The pretraining and fine-tuning data source is indicated by the respective DNA type, and `--' signifies the absence of training for that type. We also indicate the most specific taxon subset. For the baseline CNN encoder, we report the original paper result (left) and our reproduction (right). Numbers in {boldface} indicate the \first{best result} across each task, and {underlined} \runnerup{second best}.}\label{bzsl-result-table}
  \setlength{\tabcolsep}{10pt} % Default value: 6pt
  \centering
  \iftoggle{arxiv}{\small}{\scriptsize}
  \resizebox{\ifdim\textwidth>\width\width\else\textwidth\fi}{!}{
  %\begin{tabular*}{\textwidth}{@{\extracolsep{\fill}}lllrrr@{\extracolsep{\fill}}}
  \begin{tabular}{lllrrr}
    \toprule
    & \multicolumn{2}{c}{Data sources} & \multicolumn{3}{c}{Species-level acc (\%)} \\
    \cmidrule(r){2-3} \cmidrule(l){4-6}
    Model & SSL pretraining & Fine-tuning & Seen     & Unseen    & Harmonic Mean \\
    \midrule
    CNN encoder    & -- & Insect  & 38.3 / \runnerup{39.4} &  \first{20.8} / \runnerup{18.9} &  \first{27.0} / \first{25.5} \\
    DNABERT       & Human & --   & 35.0 &  10.3 &  16.0 \\
    DNABERT       & Human & Insect & \first{39.8} &  10.4 &  16.5 \\
    DNABERT-2     & Multi-species & --  & 36.2 &  10.4 &  16.2 \\
    DNABERT-2     & Multi-species & Insect & 30.8 &   8.6 &  13.4 \\
    \midrule
    BarcodeBERT (ours)     & Invertebrates & -- & 31.6 & \first{20.0} & \runnerup{24.5} \\
    BarcodeBERT (ours)      & Invertebrates & Insect & \runnerup{38.8} & 15.3 & 22.0 \\
    %BarcodeBERT (ours)     & BIOSCAN-5M & 32.5 &  & 22.4 & 26.5 \\
    %BarcodeBERT (ours)      & BIOSCAN-5M & Insect & 34.5 & 21.6 & 26.5 \\
    \bottomrule
  %\end{tabular*}
  \end{tabular}
  }
  \iftoggle{arxiv}{\footnotesize}{\vspace{-1mm}}
\end{table*}

\subsection{Ablation Studies}
In our main results, we demonstrated the utility of self-supervised pretraining BarcodeBERT to enable our model to generalize to open-world tasks. In this section, we study the impact of the different components involved in pretraining. We use the terminology {\it context tokens} for tokens that are left unchanged to provide context to the model during pretraining and the terminology {\it substitution tokens} for tokens that will be changed for the masked language modelling task. We consider different strategies to calculate the loss of each group separately. The loss associated with predicting contextual tokens is referred to as the ``context component'' of the loss\footnote{Although predictions in the foundation model literature are typically restricted to substitution tokens, we extend this to include context token predictions, maintaining the terminology to explore their utility in input reconstruction similar to autoencoders.}, while the loss related to predicting substitution tokens is the ``substitution component''. By assigning different weights to these two loss components, we sought to observe how these adjustments would affect both training and evaluation. In particular, we define $w_s$ as the penalty weight given to the substitution component of the loss. Note that $1\!-\!w_s$ is always the weight of the context component of the loss. 

\subsubsection{Substitution token rate}
To examine how varying the substitution token ratio ($r_s$) affects performance, we tested several ratios, keeping the model architecture (4 attention heads, 4 layers), tokenization ($k\!=\!4$) and substitution penalty weight ($w_s\!=\!1$) constant. \autoref{tab:substitution_ratio} shows that species-level classification performance remains consistently high across substitution rates, peaking at 99.67\% accuracy with 45\% and 50\% substitution tokens. Linear probe results align closely, reaching the highest accuracy of 99.02\% at the 50\% substitution rate. For genus-level 1-NN probing of unseen species, the 50\% substitution rate yields the best accuracy at 78.47\%, suggesting that this rate provides a balance that strengthens the model’s ability to generalize to new taxa. Lower substitution rates show slightly reduced generalization, while a 60\% rate begins to degrade performance, indicating that 50\% is the optimal value for $r_s$.

\subsubsection{Weight of the substitution component of the loss}\label{loss_weight_experiments}
Building on the fact that predicting context tokens is inherently easier than predicting substitution tokens for LLMs, we investigated how adjusting the penalty weights between these two tasks affects the performance of the model. For this purpose, we experimented with different penalty weights assigned to the substitution component of the loss ($w_s$). 
\autoref{tab:loss-weight-experiment} provides the accuracy for genus-level 1-NN probing of unseen species across different values for $w_s$ in combination with four $k$-mer sizes (2, 4, 6, and 8) and Byte-Pair Encoding (BPE) tokenizer obtained from \mbox{DNABERT-2}. Alternative BPE tokenizers that specifically fit our data are investigated later in this section.  We kept the architecture (4 layers, 4 attention heads) and substitution token rate ($r_s\!=\!50\%$) constant. As shown in \autoref{tab:loss-weight-experiment}, the optimal performance across all $k$-mer sizes was achieved with a $w_s$ of 1.0, where the highest accuracy, 78.47\%, was observed with $k\!=\!4$.

\begin{table}[tbh]
\caption{Classification accuracy over the different substitution token ratios $r_s$, while keeping constant all of the model architecture (4-4), the value of $k\!=\!4$ during tokenization and the penalty weight for the substitution component of the loss ($w_s\!=\!1$). Numbers in \first{boldface} indicate the \first{highest accuracies} and the {italic} value shows the \selected{selected} shows the selected optimal parameter.}
\iftoggle{arxiv}{\small}{\scriptsize}
\centering
\begin{tabular}{@{}cccc@{}}
\toprule
  % \multirow[b]{2}{*}[-1.5mm]{\shortstack[l]{Proportion of  \\ substitution \\ tokens (\%)}}
  \multirowcell{2}{ Substitution  \\ token\\ ratio (\%)}
  & \multicolumn{2}{l}{\makecell{Species-level acc (\%)\\ of seen species}} &  \multicolumn{1}{l}{\makecell{Genus-level acc (\%)\\ of unseen species}} \\ 
 \cmidrule(l){2-3} \cmidrule(l){4-4}
     & Fine-tuned & Linear probe & 1-NN probe \\ \midrule
15 & 98.95 & 98.95  & 75.15 \\
30 & 98.79  & 98.79 & 74.24 \\
45 & {\bf 99.67} & 98.54 & 74.42 \\
\selected{50} & {\bf 99.67} & {\bf 99.02}  & {\bf 78.47} \\
60 & 99.62 & 98.45 & 77.56 \\
\bottomrule
\end{tabular}% 
\label{tab:substitution_ratio}
\end{table}

\begin{table}[tbh]
\caption{Genus-level accuracy for 1-NN probing of unseen species with varying penalty weight assigned to the substitution component of the loss ($w_s$). The model architecture remains fixed (4-4), with substitution token ratio $r_s$ = 50\%. Two tokenizers were tested: a $k$-mer tokenizer with $k$-mer sizes of 2, 4, 6, and 8, and a BPE tokenizer used in DNABERT-2 with a fixed vocabulary size of 4096. Note that the weight of the context component of the loss always equals $1-w_s$. The number in boldface indicates the overall \first{best accuracy}, underlined the \runnerup{best per tokenizer}, and {italic} the \selected{selected} optimal parameter.}
\centering
\label{tab:loss-weight-experiment}
\iftoggle{arxiv}{\small}{\scriptsize}
\setlength{\tabcolsep}{3pt} % Adjust column padding
\renewcommand{\arraystretch}{1.1} % Adjust row padding
\begin{tabular}
{@{\extracolsep{\fill}}cccccc@{\extracolsep{\fill}}}
\toprule
& \multicolumn{5}{c}{Genus-level acc (\%) of} \\
 & \multicolumn{5}{c}{unseen species with $1$-NN probe} \\
\cmidrule(l){2-6}
\multicolumn{1}{c}{Loss weight ($w_s$)} & $k\!=\!2$ &$k\!=\!4$ &$k\!=\!6$ &$k\!=\!8$ & BPE\\
% \midrule
% 0.8 & 0.2 & 65.07 & 73.86 & 74.89 & 71.85 & 69.21  \\
% 0.5 & 0.5 & 67.13 & 75.26 & 73.95 & 74.07 & 67.50  \\
% 0.2 & 0.8 & 62.55 & 75.68 & 73.65 & 67.85 & 67.92 \\
% 0.0 & 1.0 & 76.39 & \textbf{77.07} & 75.43 & 76.18 &  69.61\\
% \bottomrule

\midrule
0.2 & 64.18 & 76.06 & 75.15 & 71.15 & \runnerup{70.57}  \\
0.5 & 66.47 & 74.98 & \runnerup{76.62} & 71.22 & 70.34  \\
0.8 & 68.84 & 76.71 & 74.66 & 73.33 & 69.40 \\
0.9 &  69.51     & 77.16 &    76.06   &  72.23     & 67.48       \\
\selected{1.0} & \runnerup{76.92} & \textbf{78.47} & 75.74 & \runnerup{75.62} & 69.85\\
\bottomrule

\end{tabular}
\iftoggle{arxiv}{}{\vspace{-2mm}}
\end{table}

Our experiments indicate that focusing the loss penalty on the harder task of predicting substitution tokens, while not penalizing the easier task of predicting context tokens, yields the best accuracy. This aligns with observations in other foundation models, such as BERT and DNABERT (see \iftoggle{arxiv}{\appref{a:masking}}{Supplement~C} for more details).

\subsubsection{Tokenization strategies}
\begin{table*}[pht]\centering
\caption{Genus-level accuracy for unseen species with different tokenizers, various model sizes, fixed weight for the substitution component of the loss function ($w_s\!=\!1$) and a substitution token ratio $r_s\!=\! 50\%$. Two types of tokenizers were tested: a $k$-mer tokenizer with $k$-mer sizes of 2, 4, 6, and 8, and five different versions of BPE tokenizers. BPE tokenizer from DNABERT-2 has a fixed vocabulary size of 4096, while BPE tokenizers trained on our dataset have vocabulary sizes ($v$) of 4096, 1024, 256, and 128. Numbers in boldface indicate the \first{best} result across each architecture.}
\label{tab:bpe-experiment}
\iftoggle{arxiv}{\small}{\scriptsize}
\setlength{\tabcolsep}{1pt} % Adjust column padding
\renewcommand{\arraystretch}{1.1} % Adjust row padding
\begin{tabular*}{\textwidth}{@{\extracolsep{\fill}}lccccccccc@{\extracolsep{\fill}}}
    \toprule
    & \multicolumn{4}{c}{$k$-mer tokenizer} & DNABERT-2 BPE & \multicolumn{4}{c}{BarcodeBERT BPE} \\
    \cmidrule(l){2-5} \cmidrule(l){6-6} \cmidrule(l){7-10}
    Model size & $k\!=\!2$ & $k\!=\!4$ & $k\!=\!6$ & $k\!=\!8$ & $v\!=\!4096$ & $v\!=\!4096$ & $v\!=\!1024$ & $v\!=\!256$ & $v\!=\!128$ \\ 
    % \cmidrule(l){3-5} \cmidrule(l){6-6} \cmidrule(l){7-10}
    \midrule

    4 layers, 4 heads & 76.92 & \textbf{78.47} & 75.74 & 75.62 & 69.85 & 66.88 & 68.58 & 66.57 & 63.42 \\
    6 layers, 6 heads & 71.46 & \textbf{76.95} & 76.04 & 76.60 & 70.17 & 67.30 & 66.95 & 63.49 & 60.61 \\
    12 layers, 12 heads & 74.71 & 70.17 & 70.80 & \textbf{75.81} & 68.68 & 67.79 & 62.39 & 56.94 & 54.09 \\
    \bottomrule
\end{tabular*}
\iftoggle{arxiv}{}{\vspace{-1mm}}
\end{table*}
We evaluated the BPE tokenizer, which generates variable-length tokens based on character co-occurrence frequencies \citep{sennrich2015neural}. 
% Derived from a data compression algorithm \citep{gage1994new}, 
BPE was designed for subword tokenization to overcome fixed vocabulary \citep{sennrich2015neural} by compressing sequences for efficient representation \citep{Gall2019InvestigatingTE}. Unlike overlapping $k$-mers, BPE has an advantage for masked DNA language models \citep{zhou2023dnabert2} as it avoids information leaks from adjacent masked tokens. 
We used \mbox{DNABERT-2}'s BPE tokenizer, trained on 2.75 billion nucleotide bases from the human nuclear genome and 32.49 billion bases from 135 species across various kingdoms. We also trained custom BPE tokenizers with varying vocabulary sizes on our DNA barcode dataset and evaluated across various training scenarios on the genus-level 1-NN probing task. Based on results in \autoref{tab:loss-weight-experiment}, we selected a loss weight of 1.0 for best performance. Our setup used a 50\% substitution token ratio ($r_s$). Full results across configurations are shown in \autoref{tab:bpe-experiment}.  

\noindent\mypara{Model size} We report genus-level 1-NN probing accuracy on the unseen data for DNABERT-2 BPE and BarcodeBERT BPE tokenizers across three model sizes. Models using \mbox{DNABERT-2's} BPE showed no significant accuracy changes across different sizes. However, for BarcodeBERT BPE, we see that increasing the model size reduces the accuracy across different vocabulary sizes, possibly due to overfitting. For additional experiments, see Supplementary Materials~D.3.

\noindent\mypara{Vocabulary size} We trained new BPE tokenizers with varying vocabulary sizes ($v$) to evaluate their impact on BarcodeBERT BPE performance. Unlike $k$-mer tokenizers, BPE sequence lengths vary with input composition and require padding or truncation to a maximum length before tokenization. 
Since smaller vocabularies ($v$) produce longer sequences (see Supplement~B.1), we set maximum lengths of 128 for $v\!\in\!\{4096, 1024\}$ and 256 for $v\!\in\!\{256, 128\}$.  
Reducing the vocabulary size from 4096 to 128 consistently decreases accuracy across all model sizes.

\noindent\mypara{Comparing $k$-mer with BPE} We find that $k$-mer tokenizers outperform BPE tokenizers in all model configurations. We hypothesize that this is likely due to three reasons. First, DNA barcode sequences are too short to benefit from the BPE compression. Second, BPE is sensitive to minor variations such as single-character substitutions, which is unsuitable for DNA datasets with single-nucleotide mutations in arbitrary positions. In other words, 
BPE tokenization varies greatly for similar sequences with small Hamming distances, while $k$-mer tokenization remains consistent, as a single-nucleotide substitution affects only one token (see Supplementary Figure~S9). % Supp Fig S9
Third, although BPE handles small sequence alignment shifts better than $k$-mer tokenizers, this can be mitigated in $k$-mers with data augmentation using random offsets during pretraining (see Supplementary Table S6).

% \vspace{-0.1cm}
\section{Discussion}
Our results demonstrate that pretraining masked language models on DNA barcode data, as exemplified by BarcodeBERT, is highly effective for arthropod species identification. Despite only ~35\% of the pretraining data being labelled at the species rank, the model learns latent features that purely supervised approaches cannot capture. BarcodeBERT performs well on key biodiversity tasks, such as taxonomic classification, clustering, and similarity searches, benefiting from efficient hardware acceleration that enables scaling for large datasets and achieving $55\times$ faster species-level classification than alignment-based methods.
By systematically evaluating tokenization and masking strategies, we also provide actionable insights for the pretraining of  DNA-specific foundation models. 
% Additionally, BarcodeBERT’s efficient use of hardware acceleration enables it to scale effectively for large datasets while maintaining inference times that are faster than alignment-based approaches.

Despite its strengths, BarcodeBERT has some limitations. Its training data may have taxonomic and geographical biases as the model is trained exclusively on invertebrate species from Canada, potentially limiting its applicability in global studies. The BOLD dataset, comprising more than 16 million DNA barcodes from a wide geographical distribution \citep{RatnasinghamBOLD2007}, represents untapped data that could address such biases. Future work should incorporate more diverse datasets to develop robust, globally scalable models for taxonomic classification. Our methodology and findings should be broadly applicable to barcode regions for other kingdoms, such as the ITS region for fungi \citep{doi:fugal_ITS}, however, the method requires validation beyond the COI barcode region used for Animalia.

Lastly, while longer genomic sequences could offer deeper insights for specialized phylogenetic analyses, the quadratic time complexity of transformer models limits their applicability to such sequences. Future work should include more computationally efficient architectures such as structured state space models, which scale sub-quadratically with sequence length \citep{Gu2022-ng}.

% \vspace{-0.1cm}
\section{Conclusions}
BarcodeBERT leverages 1 million DNA barcodes with partial taxonomic annotations to outperform state-of-the-art foundation models in genus-level and species-level classification tasks. Notably, BarcodeBERT matches the high accuracy of the alignment-based classification tool 
BLAST in species classification, while being 55 times faster and more scalable.   
In addition, our extensive analysis of pretraining strategies provides actionable insights for building customized DNA language models for large-scale taxonomic classification.

Overall, BarcodeBERT's performance demonstrates how transformer-based architectures can be successfully customized to overcome the challenges of genomic biodiversity data for effective DNA barcode identification and classification. Lastly, not being limited to a specific dataset or barcode region, our model is highly amenable to future applications, to global datasets or barcode regions in other kingdoms of life.

% \vspace{-0.2cm}
\FloatBarrier
\section{Competing interests}
No competing interest is declared.

% \vspace{-0.2cm}
\section{Author contributions statement}
PMA curated the data, and partitioned it with assistance from GWT, DS, MS. PMA, SCL, MS, and NS implemented BarcodeBERT. PMA ran the DNA baseline experiments. MS conducted the masking and loss penalty ablation studies. NS conducted tokenization ablation studies. ZMG and ATW conducted the multimodal retrieval learning experiments with PMA's assistance. PMA, MS, NS, LK, and IZ authored the manuscript text and figures. SCL, GWT, AXC, DS, LK, and JBH provided guidance on experimental design and edited the manuscript. All authors reviewed the manuscript.

% \vspace{-0.2cm}
\section{Acknowledgments}
We acknowledge the support of the Government of Canada’s New Frontiers in Research Fund [NFRFT-2020-00073]. This research was supported, in part, by the Province of Ontario and the Government of Canada through the Canadian Institute for Advanced Research (CIFAR), and \hreffoot{https://vectorinstitute.ai/partnerships/current-partners/}{companies sponsoring} the Vector Institute. %(\url{http://www.vectorinstitute.ai/#partners}). 
GWT is supported by the Natural Sciences and Engineering Research Council of Canada (NSERC), the Canada Research Chairs program, and the Canada CIFAR AI Chairs program. LK is supported by NSERC Discovery Grant RGPIN-2023-03663. DS is supported by the Canada First Research Excellence Fund through the University of Guelph’s ``Food From Thought'' program [Project 000054].% The funders had no role in the preparation of the manuscript.
%We acknowledge the support of the Natural Sciences and Engineering Research Council of Canada (NSERC), [funding reference number RGPIN-2023-03663]. This research was enabled, in part, by support provided by Compute Canada, the Canada First Research Excellence Fund to the University of Guelph’s “Food From Thought” research program (Project 000054), and the Government of Canada’s New Frontiers in Research Fund (NFRF) [NFRFT-2020-00073]. Resources used in preparing this research were provided, in part, by the Province of Ontario, the Government of Canada through the Canadian Institute for Advanced Research (CIFAR), and companies sponsoring the Vector Institute (\url{http://www.vectorinstitute.ai/\#partners}). The funders had no role in the preparation of the manuscript.

\bibliographystyle{icml2024} %plain
\bibliography{reference}

\appendix
\section{Dataset Description}
\label{a:dataset}

After pre-processing (see \iftoggle{arxiv}{\autoref{s:data-preprocessing}}{\textit{``Data pre-processing''} in the main text}), the final dataset contained 965,289  samples. The dataset was divided into three main partitions: \textit{Pretrain}, \textit{Seen}, and \textit{Unseen}. The \textit{Seen} partition was subdivided into supervised training, validation, and testing subsets. Each partition serves distinct experimental purposes, with the \textit{Unseen} partition designed to mimic real-world scenarios where models encounter sequences from previously unobserved species.
Our preprocessed version of the dataset is available to download at \url{https://huggingface.co/datasets/bioscan-ml/CanadianInvertebrates-ML}.

\autoref{tab:S1} summarizes the composition of each partition, detailing the number of unique records across taxonomic categories, from phylum to species. It also includes the number of unique Barcode Index Numbers (BINs); each BIN serves as a species-level proxy, created by clustering similar DNA barcode sequences into a molecular taxonomic unit. The \textit{Pretrain} partition contains 893,744 sequences from 15 phyla, while the \textit{Seen} and \textit{Unseen} partitions are more specific, containing sequences primarily from phylum Arthropoda. The \textit{Unseen} partition contains 4,278 barcodes from 1,826 species absent from the training and validation subsets.

\begin{table}[h!]
\caption{Number of unique records across taxonomic categories in each partition.}\label{tab:S1}
    \centering
    %\resizebox{\textwidth}{!}{
    \iftoggle{arxiv}{\small}{\footnotesize}
\begin{tabular}{lrrrrrrrr}
\toprule
Partition & Phylum & Class & Order & Family & Genus & Species & DNA barcode & BIN \\
\midrule
 Pretrain   &15 &49 &176 &1,188 &6,679 &14,794 &893,744 &62,489 \\
 Test       & 1 & 1 & 13 &  161 &  500 & 1,653 & 13,460 & 2,125 \\
 Train      & 1 & 1 & 13 &  161 &  500 & 1,653 & 47,086 & 2,430 \\
 Validation & 1 & 1 & 13 &  161 &  500 & 1,653 &  6,721 & 1,954 \\
 Unseen     & 1 & 1 & 13 &  161 &  500 & 1,826 &  4,278 & 1,885 \\
\bottomrule
\end{tabular}
\end{table}

To verify the overlap between partitions, Tables \ref{tab:S2} and \ref{tab:S3} provide pairwise comparisons of shared species and genera, respectively. As expected, there is no overlap in species between the \textit{Unseen} partition and other subsets, ensuring the appropriateness of the \textit{Unseen} partition for evaluating generalization. That being said, some genera are shared across partitions, reflecting the hierarchical taxonomic structure of the data. All the subsets in the \textit{Seen} partition share all 500 genera that are also present in the \textit{Unseen} partition. The \textit{Pretrain} partition shares 467 of these genera with each of the others.
\begin{table}[htb]
\caption{Pairwise comparison of species overlap between partitions, with each cell representing the number of species shared between two partitions. The total number of unique species within a partition is given along the diagonal.}\label{tab:S2}
    \centering
    %\resizebox{\textwidth}{!}{
    \iftoggle{arxiv}{\small}{\footnotesize}
\begin{tabular}{lrrrrr}
\toprule
 & Pretrain & Unseen & Train & Test & Validation \\
\midrule
Pretrain    & 14,794 &    0 &  809 &  809 &  809 \\
Unseen      &      0 & 1826 &    0 &    0 &    0 \\
Train       &    809 &    0 & 1653 & 1653 & 1653 \\
Test        &    809 &    0 & 1653 & 1653 & 1653 \\
Validation  &    809 &    0 & 1653 & 1653 & 1653 \\
\bottomrule
\end{tabular}
\end{table}

\begin{table}[htb]
\caption{Pairwise comparison of genus overlap between partitions, with each cell representing the number of genera shared between two partitions. 
 The total number of unique genera within a partition is given along the diagonal.}\label{tab:S3}
    \centering
    %\resizebox{\textwidth}{!}{
    \iftoggle{arxiv}{\small}{\footnotesize}
\begin{tabular}{lrrrrr}
\toprule
 & Pretrain & Unseen & Train & Test & Validation \\
\midrule
Pretrain & 6679 & 467 & 467 & 467 & 467 \\
Unseen & 467 & 500 & 500 & 500 & 500 \\
Train & 467 & 500 & 500 & 500 & 500 \\
Test & 467 & 500 & 500 & 500 & 500 \\
Validation & 467 & 500 & 500 & 500 & 500 \\
\bottomrule
\end{tabular}
\end{table}

\autoref{tab:S4} details the number of unique records across taxonomic ranks for each phylum in the \textit{Pretrain} partition. Arthropoda dominates this partition, accounting for over 95\% of the sequences, followed by smaller contributions from other phyla like Mollusca and Annelida. \autoref{tab:S5} highlights the percentage of labelled records for each phylum, showing significant gaps in taxonomic annotations at the species and genus levels, particularly for Arthropoda, where less than 40\% of the sequences are labelled at the species level.

\begin{table}[htb]
\caption{The distribution of barcode sequences across taxonomic ranks for each phylum used in the \textit{Pretrain} partition.}\label{tab:S4}
    \centering
    %\resizebox{\textwidth}{!}{
    \iftoggle{arxiv}{\small}{\footnotesize}
\begin{tabular}{lrrrrrr}
\toprule
Phylum & Class & Order & Family & Genus & Species & BIN \\
\midrule
Annelida & 2 & 16 & 48 & 150 & 329 & 516 \\
Arthropoda & 14 & 67 & 929 & 6,211 & 13,991 & 61,328 \\
Brachiopoda & 1 & 2 & 2 & 2 & 2 & 2 \\
Bryozoa & 3 & 3 & 3 & 2 & 2 & 4 \\
Chordata & 5 & 18 & 37 & 67 & 89 & 102 \\
Cnidaria & 4 & 10 & 24 & 25 & 24 & 46 \\
Echinodermata & 5 & 17 & 26 & 43 & 74 & 79 \\
Hemichordata & 1 & 1 & 1 & 2 & 1 & 2 \\
Mollusca & 6 & 30 & 97 & 162 & 271 & 372 \\
Nematoda & 2 & 5 & 10 & 5 & 2 & 8 \\
Nemertea & 3 & 2 & 5 & 5 & 5 & 22 \\
Platyhelminthes & 0 & 0 & 0 & 0 & 0 & 1 \\
Porifera & 1 & 3 & 4 & 4 & 3 & 5 \\
Priapulida & 1 & 1 & 1 & 1 & 1 & 1 \\
Tardigrada & 1 & 1 & 1 & 0 & 0 & 1 \\
\bottomrule
\end{tabular}
\end{table} 

\begin{table}[htb]
\caption{Percentage of labelled records across taxonomic ranks for each phylum in the \textit{Pretrain} partition.}\label{tab:S5}
    \centering
    %\resizebox{\textwidth}{!}{
    \iftoggle{arxiv}{\small}{\footnotesize}
\begin{tabular}{lrrrrrr}
\toprule
Phylum & Class & Order & Family & Genus & Species & BIN \\
\midrule
Annelida & 100\phantom{.00} & 96.43 & 99.81 & 87.44 & 80.35 & 97.95 \\
Arthropoda & 100\phantom{.00} & 100\phantom{.00} & 99.97 & 66.91 & 35.05 & 99.37 \\
Brachiopoda & 100\phantom{.00} & 100\phantom{.00} & 100\phantom{.00} & 100\phantom{.00} & 90{.00} & 90{.00} \\
Bryozoa & 80{.00} & 80{.00} & 80{.00} & 40{.00} & 40{.00} & 100\phantom{.00} \\
Chordata & 100\phantom{.00} & 99.65 & 98.62 & 98.27 & 94.81 & 96.89 \\
Cnidaria & 100\phantom{.00} & 98.21 & 95.54 & 84.82 & 54.46 & 99.11 \\
Echinodermata & 100\phantom{.00} & 100\phantom{.00} & 100\phantom{.00} & 100\phantom{.00} & 100\phantom{.00} & 100\phantom{.00} \\
Hemichordata & 100\phantom{.00} & 100\phantom{.00} & 100\phantom{.00} & 100\phantom{.00} & 25{.00} & 100\phantom{.00} \\
Mollusca & 99.95 & 89.02 & 99.69 & 98.43 & 90.22 & 96.71 \\
Nematoda & 100\phantom{.00} & 100\phantom{.00} & 91.67 & 54.17 & 16.67 & 37.50 \\
Nemertea & 96.43 & 94.64 & 87.50 & 64.29 & 48.21 & 100\phantom{.00} \\
Platyhelminthes & 0\phantom{.00} & 0\phantom{.00} & 0\phantom{.00} & 0\phantom{.00} & 0\phantom{.00} & 100\phantom{.00} \\
Porifera & 100\phantom{.00} & 100\phantom{.00} & 100\phantom{.00} & 85.71 & 42.86 & 85.71 \\
Priapulida & 100\phantom{.00} & 100\phantom{.00} & 100\phantom{.00} & 100\phantom{.00} & 100\phantom{.00} & 100\phantom{.00} \\
Tardigrada & 100\phantom{.00} & 100\phantom{.00} & 100\phantom{.00} & 0\phantom{.00} & 0\phantom{.00} & 100\phantom{.00} \\
\bottomrule
\end{tabular}
\end{table}

\begin{figure}[!htb]
    \centering
    % First row: three subfigures
    \begin{minipage}{\textwidth}
        \centering
        \begin{subfigure}[b]{0.32\textwidth}
            \centering
            \includegraphics[width=\textwidth]{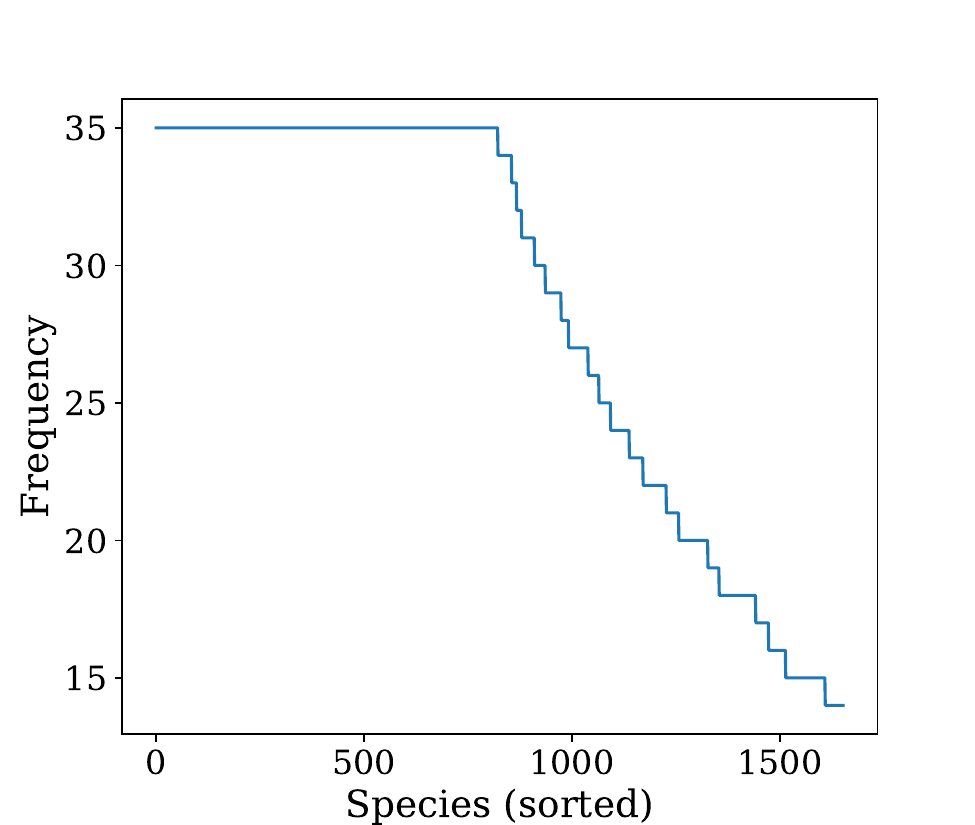}
            \caption{Train subset (\textit{Seen})}
            \label{fig:train_species_dist}
        \end{subfigure}
        \hfill
        \begin{subfigure}[b]{0.32\textwidth}
            \centering
            \includegraphics[width=\textwidth]{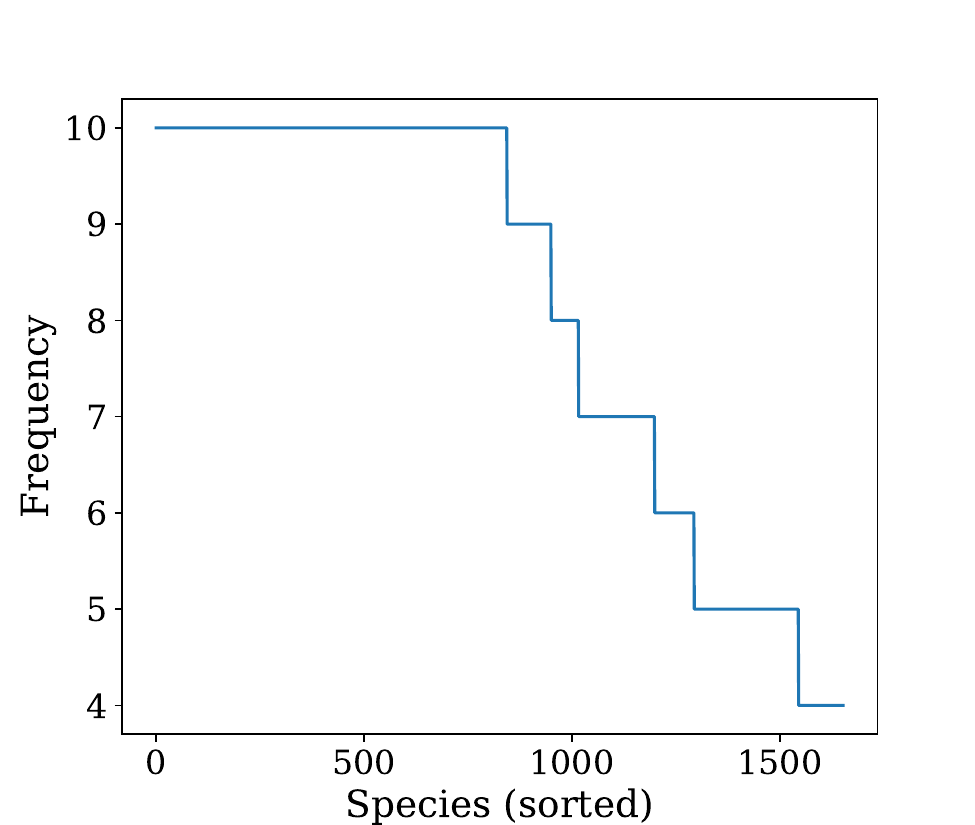}
            \caption{Test subset (\textit{Seen})}
            \label{fig:test_species_dist}
        \end{subfigure}
        \hfill
        \begin{subfigure}[b]{0.32\textwidth}
            \centering
            \includegraphics[width=\textwidth]{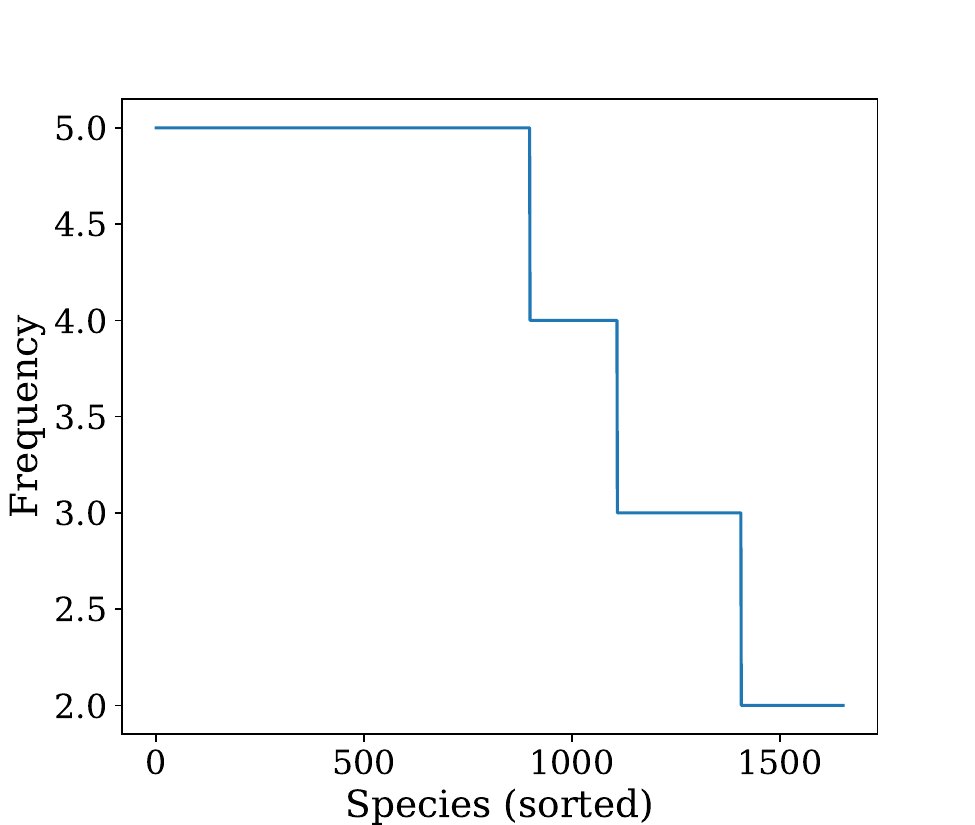}
            \caption{Validation subset (\textit{Seen})}
            \label{val_species_dist}
        \end{subfigure}
    \end{minipage}

    % Add vertical spacing between rows
    \vspace{0.5em}

    % Second row: two subfigures
    \begin{minipage}{\textwidth}
        \centering
        \hspace{3em}
        \begin{subfigure}[b]{0.32\textwidth}
            \centering
            \includegraphics[width=\textwidth]{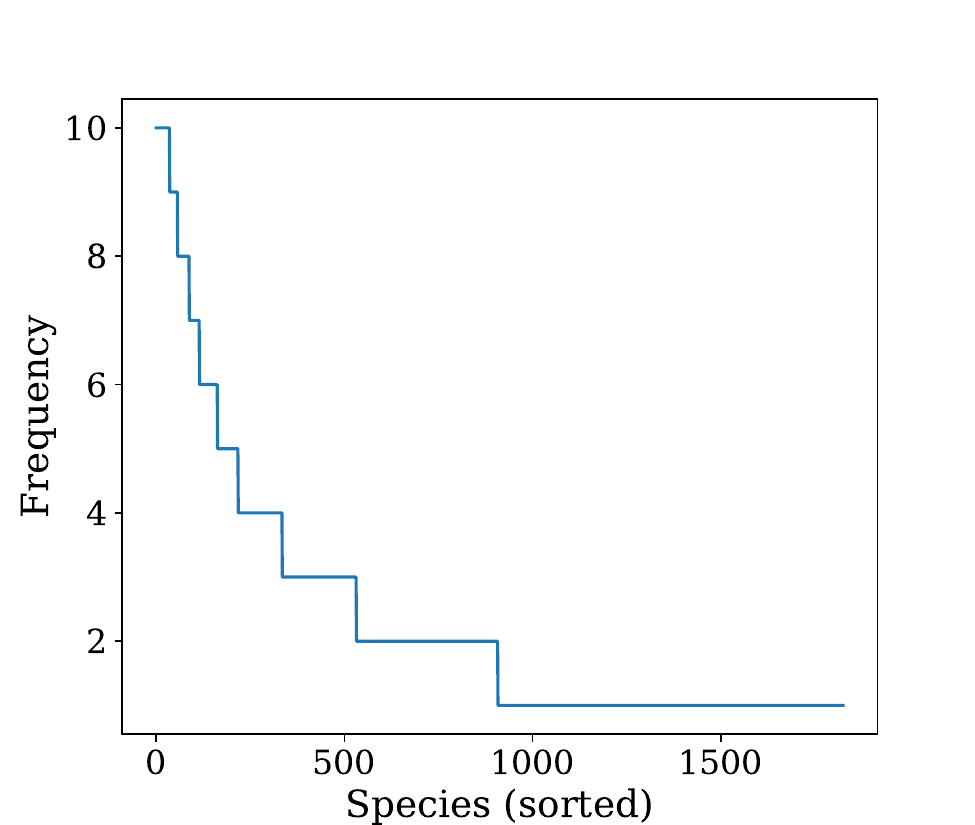}
            \caption{\textit{Unseen} partition}
            \label{fig:unseen_species_dist}
        \end{subfigure}
        \hspace{3em}
        \begin{subfigure}[b]{0.32\textwidth}
            \centering
            \includegraphics[width=\textwidth]{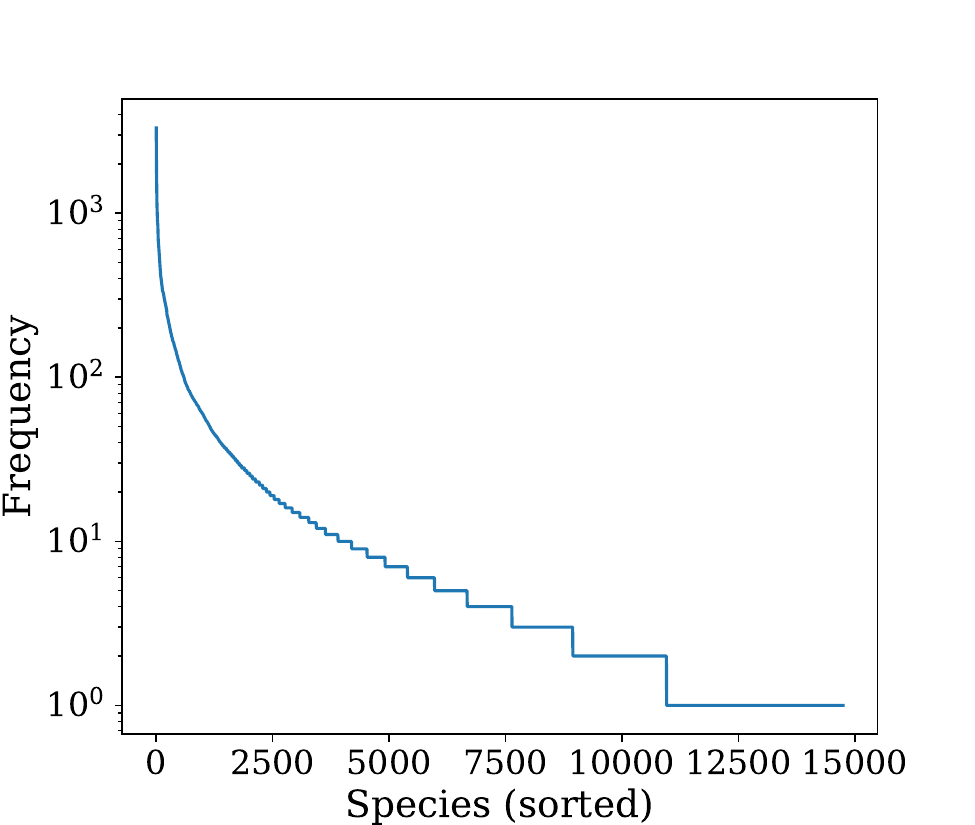}
            \caption{\textit{Pretrain} partition}
            \label{fig:pretrain_species_dist}
        \end{subfigure}
    \end{minipage}

    \caption{Frequency of species records across each partition, sorted in decreasing order.}
    \label{fig:species_dist}
\end{figure}

\begin{figure}
    \centering
    \includegraphics[width=\iftoggle{arxiv}{0.9}{}\linewidth]{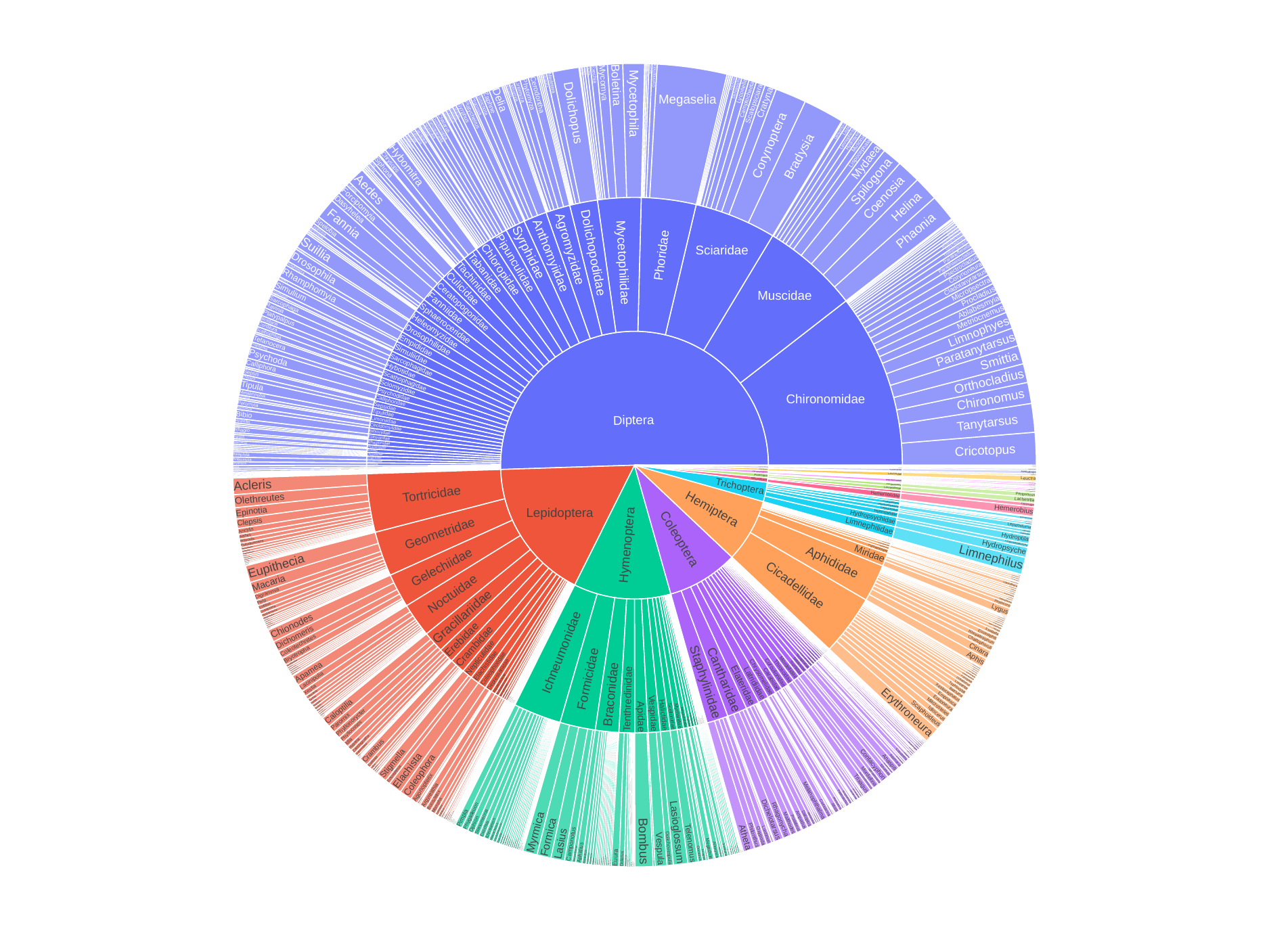}
    \caption{Sunburst plot of taxonomic distribution in the test subset of the \textit{Seen} partition, with each radial level representing a taxonomic rank from order to genus. An interactive version of this plot is available at \protect\href{https://htmlpreview.github.io/?https://github.com/bioscan-ml/BarcodeBERT/blob/main/Figures/supervised_test_taxonomic_sunburst_chart.html}{\sf Test\_Seen.html}}
    \label{fig:Sunburst_test}
\end{figure}

\begin{figure}
    \centering
    \includegraphics[width=\iftoggle{arxiv}{0.9}{}\linewidth]{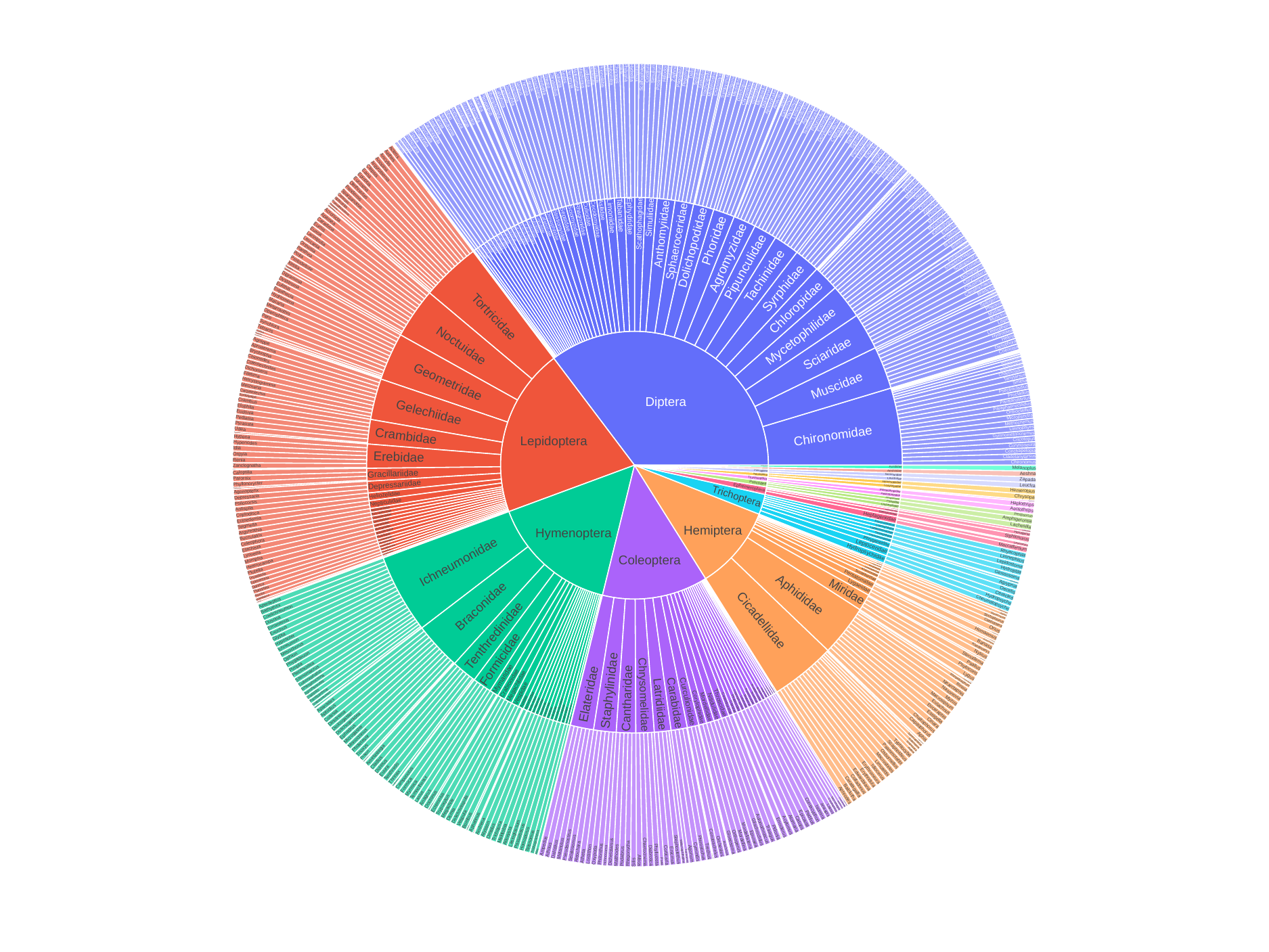}
    \caption{Sunburst plot of taxonomic distribution in the \textit{Unseen} partition, with each radial level representing a taxonomic rank from order to genus. An interactive version of this plot is available at \protect\href{https://htmlpreview.github.io/?https://github.com/bioscan-ml/BarcodeBERT/blob/main/Figures/taxonomic_sunburst_chart.html}{\sf Unseen.html}}.
    \label{fig:Sunburst_unseen}
\end{figure}

\autoref{fig:species_dist} illustrates the species frequency across partitions, sorted in decreasing order. The subsets in the \textit{Seen} partition exhibit a relatively uniform distribution, while the \textit{Unseen} partition contains a higher proportion of rare species. Figures \ref{fig:Sunburst_test} and \ref{fig:Sunburst_unseen} present sunburst plots for the supervised test subset and the \textit{Unseen} partition, visualizing the taxonomic hierarchy from order to genus. These figures highlight the differences in taxonomic composition between partitions, with the supervised test subset showing a more balanced distribution compared to the \textit{Unseen} partition, which emphasizes underrepresented genera and families.

%The partitions provide a structured dataset that supports the evaluation of models in both closed- and open-world settings. It enables a robust assessment of models' ability to classify known taxa and generalize to new species.

Lastly, we analyzed the variability of COI DNA barcodes across insect species in the \textit{test} partition using the normalized edit distance and cosine distance on the learned embeddings. Only species with more than two sequences are considered, and for each species $\mathcal{S}$, we computed both intra- and inter-species distances. Specifically, for each sequence, we calculated:
\begin{itemize}
    \item The distance to all other sequences within the same species (intra-species), and
    \item The distance to all sequences from different species (inter-species),
\end{itemize}
while ensuring distances were not double-counted. Finally, we estimated each distribution via Gaussian kernel density estimation (KDE) and computed overlap by integrating the minimum of the two KDE curves.
\definecolor{plot_blue}{HTML}{8FBBD9}
\definecolor{plot_orange}{HTML}{E4A76F}
\definecolor{plot_green}{HTML}{989D5D}
\begin{figure}[!h]
    \centering
    \includegraphics[width=0.65\linewidth]{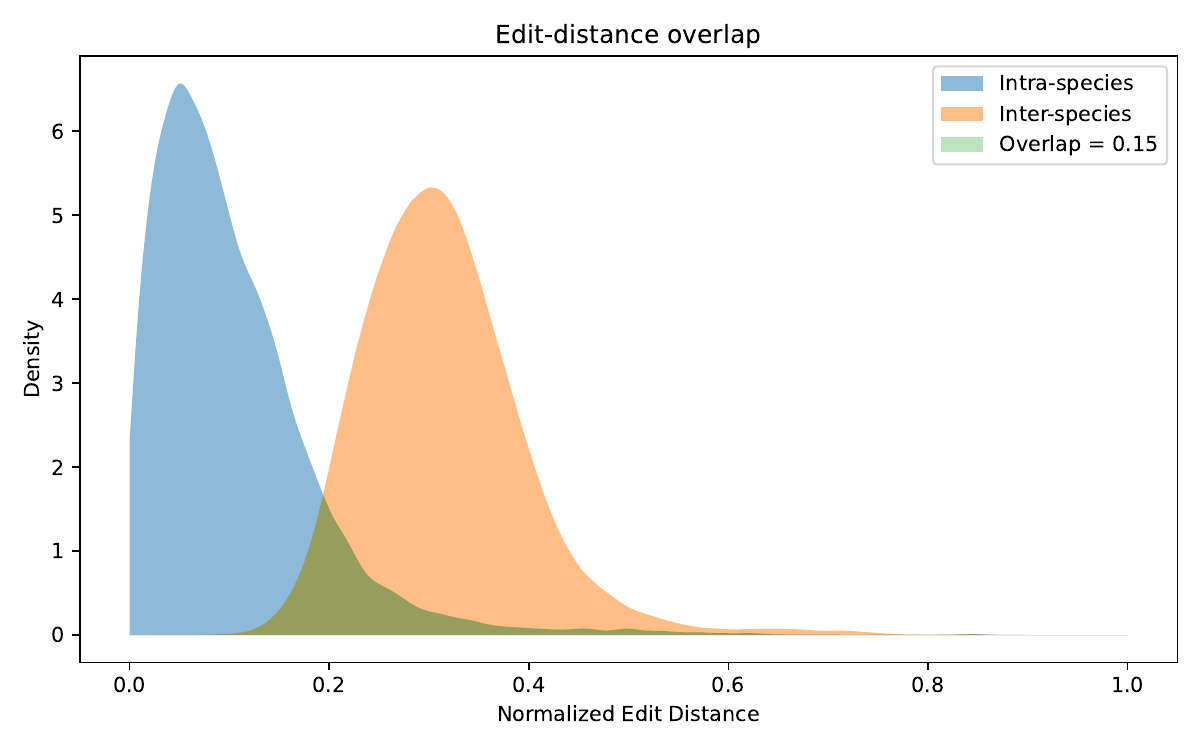}
    \caption{Distribution of intra- and inter-species distances based on \textbf{normalized edit distance} computed on the test partition. 
\textcolor{plot_blue}{\textbf{Blue}} represents intra-species distances, \textcolor{plot_orange}{\textbf{orange}} represents inter-species distances, and the area of overlap is shaded in \textcolor{plot_green}{\textbf{green}}. 
Distances were calculated using Levenshtein alignment, normalized using min-max normalization. Species with at least two sequences were included in the analysis.}
    \label{fig:edit-distance-distribution}
\end{figure}

\begin{figure}[!h]
    \centering
    \includegraphics[width=0.65\linewidth]{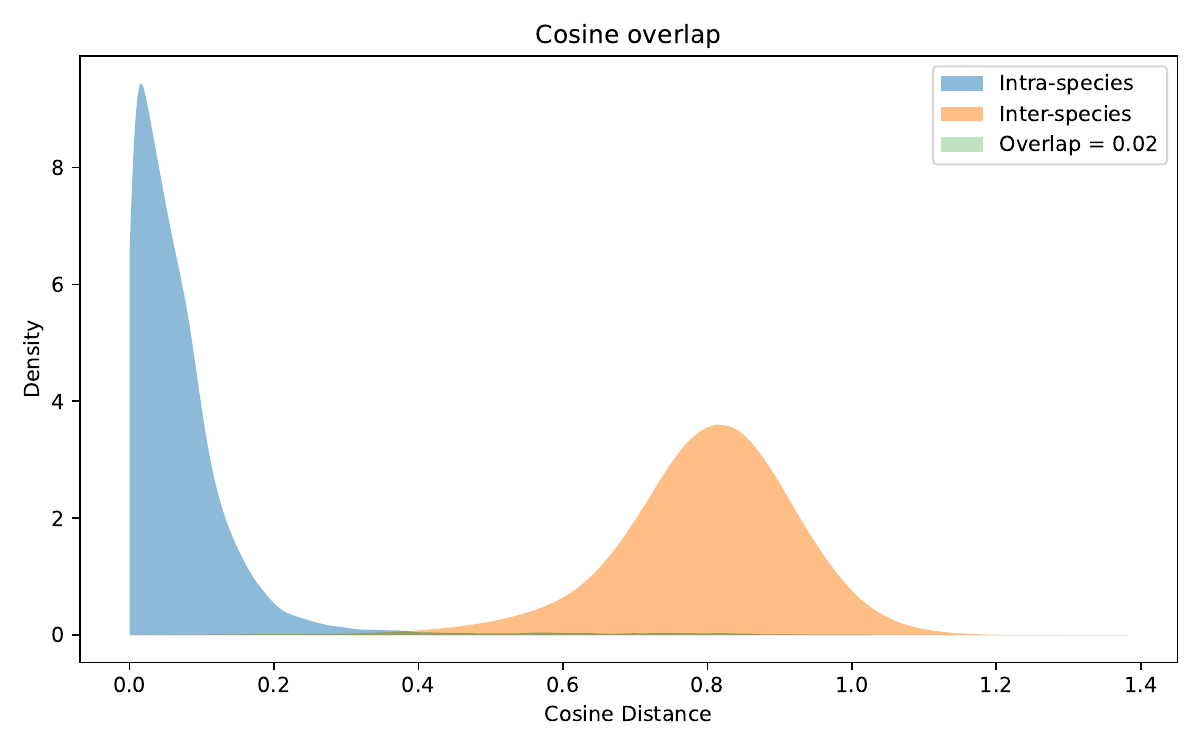}
    \caption{Distribution of intra- and inter-species distances based on cosine distance computed on GAP embeddings generated by the model. \textcolor{plot_blue}{\textbf{Blue}} represents intra-species distances, \textcolor{plot_orange}{\textbf{orange}} represents inter-species distances, and the area of overlap is shaded in \textcolor{plot_green}{\textbf{green}}. 
While cosine distances on learned embeddings capture biological structure, the separation between intra- and inter-species groups is less distinct than with alignment-based metrics. }
    \label{fig:cosine_distributions}
\end{figure}

The edit distance (\autoref{fig:edit-distance-distribution}) provides a clear separation between intra- and inter-species distributions, with minimal overlap (0.15). These distributions quantitatively capture the natural variability in COI sequences and illustrate the potential for learning taxonomic structure with deep models. This intuition is confirmed by the results obtained after estimating the distribution of the cosine distances of the learned embeddings (\autoref{fig:cosine_distributions}). Indeed, self-supervised pretraining enables BarcodeBERT to encode biologically relevant distinctions more clearly than raw alignment metrics with an overlap of 0.02 between both distributions. While these distances reflect biological structure and are useful in our $k$-NN classifier and zero-shot classification (ZSC) settings, the separation between intra- and inter-species groups should be further investigated, specially constrained to other taxonomic ranks.

\newpage
%\section{Training Details}
\section{Tokenization Strategies --- Extra results}
\label{a:tokenization}

% \subsection{Masking Strategies}
\subsection{Variation in Length of Tokenized Sequences}
Compared to the $k$-mer tokenizer, which generates a tokenized sequence proportional to the DNA sequence length, the length of the tokenized sequence produced by the BPE tokenizer is not directly determined by the nucleotide sequence length and can vary depending on the composition of the input sequence and the vocabulary size of the BPE tokenizer.

\begin{figure}[!h]
    \centering
    \includegraphics[width=0.65\linewidth]{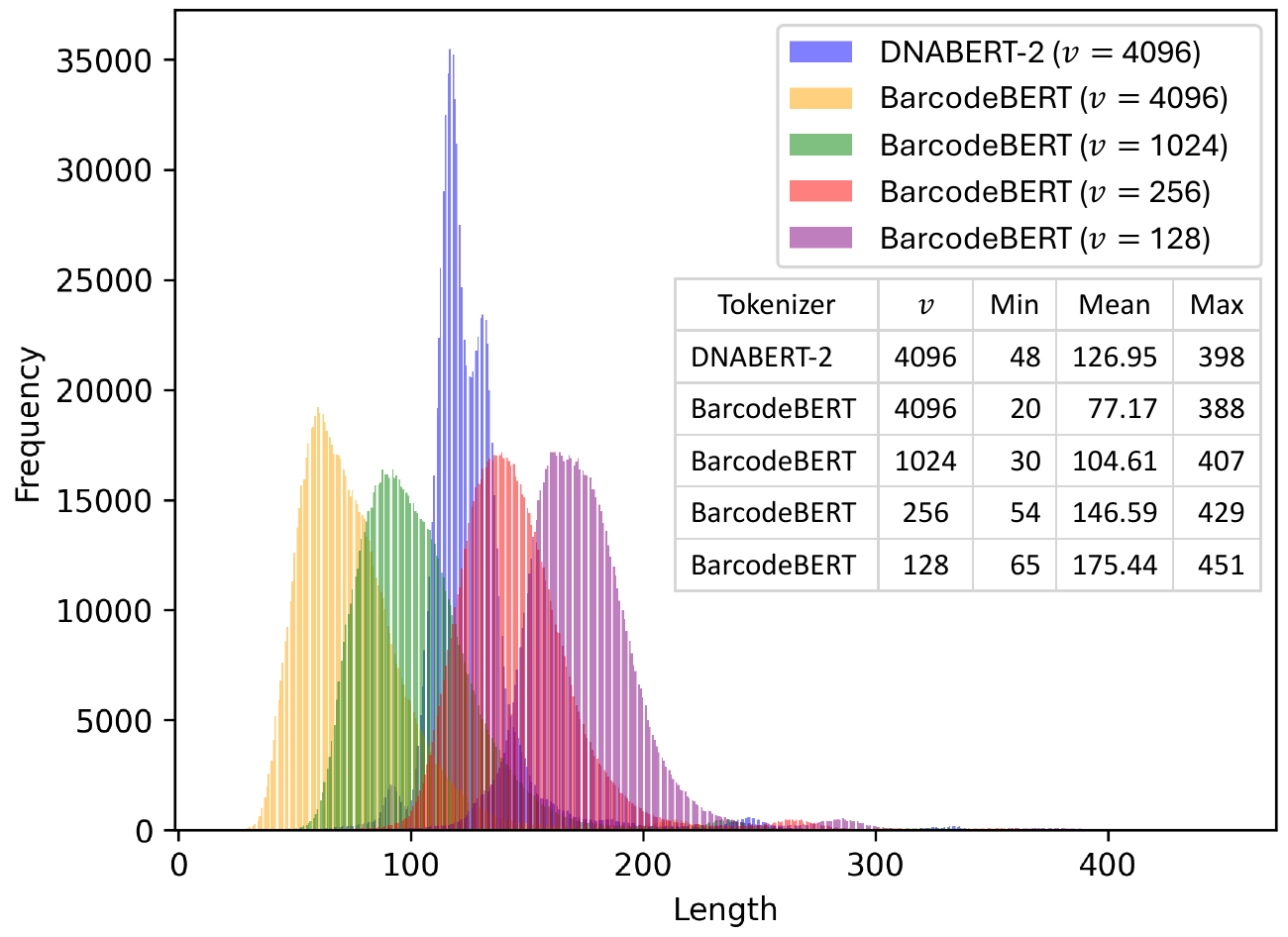}
    \caption{Distribution of tokenized sequence lengths across different BPE tokenizers, shown as histograms for the DNA barcode pretraining data. The distributions correspond to different BPE tokenizers, including DNABERT-2 and BarcodeBERT with vocabulary sizes ($v$) of 4096, 1024, 256, and 128. Smaller vocabulary sizes result in longer tokenized sequences, as reflected by the rightward shift in the distribution. Summary statistics (minimum, mean, and maximum lengths) are provided in the table inset.}
    \label{tab:bpe_tokenized_len}
\end{figure}

\autoref{tab:bpe_tokenized_len} shows the distribution of tokenized sequence lengths in DNA barcode pretraining data. Generally, BPE tokenizers with smaller vocabulary sizes tend to produce longer tokenized sequences, with the only exception being the DNABERT-2 BPE. According to \autoref{tab:bpe_tokenized_len}, although the tokenizer trained on DNABERT-2 uses a vocabulary size of 4096, it generates relatively long tokenized sequences. This is because the dataset used to train DNABERT-2 differs from the DNA barcode dataset used to create other tokenizers.

\subsection{Sensitivity of Tokenizers to DNA Sequence Variations}
%In this section we first provide an overview of string similarity measures and their relevance in comparative genomics, focusing on Hamming and edit distances, we then explore how tokenization strategies relate to the measures and hypothesize why $k$-mer tokenization performed better than BPE in our applications. \\

\begin{figure}[bth]
\centering
\begin{subfigure}[b]{0.47\textwidth}
\centering
\includegraphics[width=\textwidth]{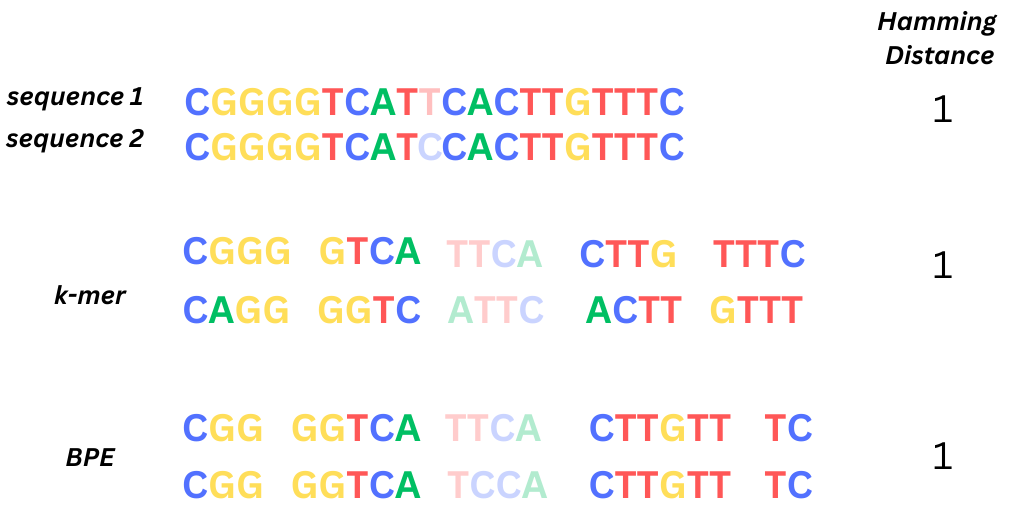}
\caption{Substitution}
\label{fig:substitution_example}
\end{subfigure}
\hfill
\begin{subfigure}[b]{0.47\textwidth}
\centering
\includegraphics[width=\textwidth]{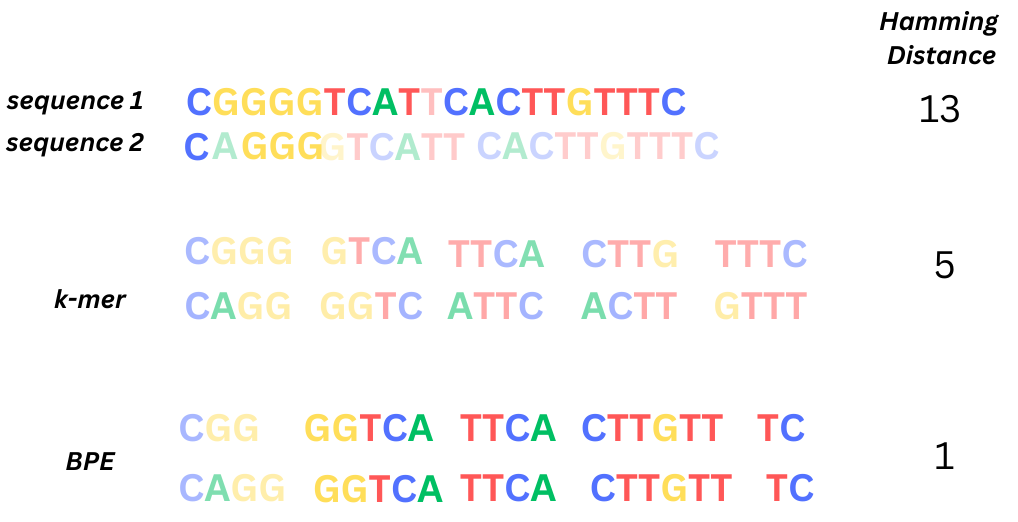}
\caption{Insertion}
\label{fig:insertion_example}
\end{subfigure}
\caption{Illustration of the differences between the Hamming distances in sequence and token space using $k$-mer and DNABERT-2 BPE tokenization strategies for two different string operations where the edit distance between a pair of sequences is equal to one.}
\label{fig:example_distances}
\end{figure}

There are two main ways to measure similarity between sequences. One is through the Hamming distance, which counts the number of mismatches between two strings, i.e., the number of positions where the strings do not share the same nucleotide. The other is the edit distance, also called the Levenshtein distance, which is defined as the minimum number of string operations (substitutions, insertions, deletions) required to transform one string into the other. In comparative genomics, the edit distance is widely used to estimate evolutionary relationships, as it corresponds to a better model of true sequence similarity. However, the Hamming distance can be calculated in linear time, even in token space with larger vocabularies. \autoref{fig:example_distances} illustrates how the Hamming distance varies with a single insertion operation versus a single substitution operation in both sequence space and token space for a pair of sequences. We include the two tokenization strategies considered in our study: $k$-mer and DNABERT-2 BPE tokenization. In both scenarios, the edit distance is equal to one, but just the BPE tokenizer can model this behaviour using the Hamming distance as a proxy.\\

\begin{figure}[bth]
\centering
\includegraphics[width=\linewidth]{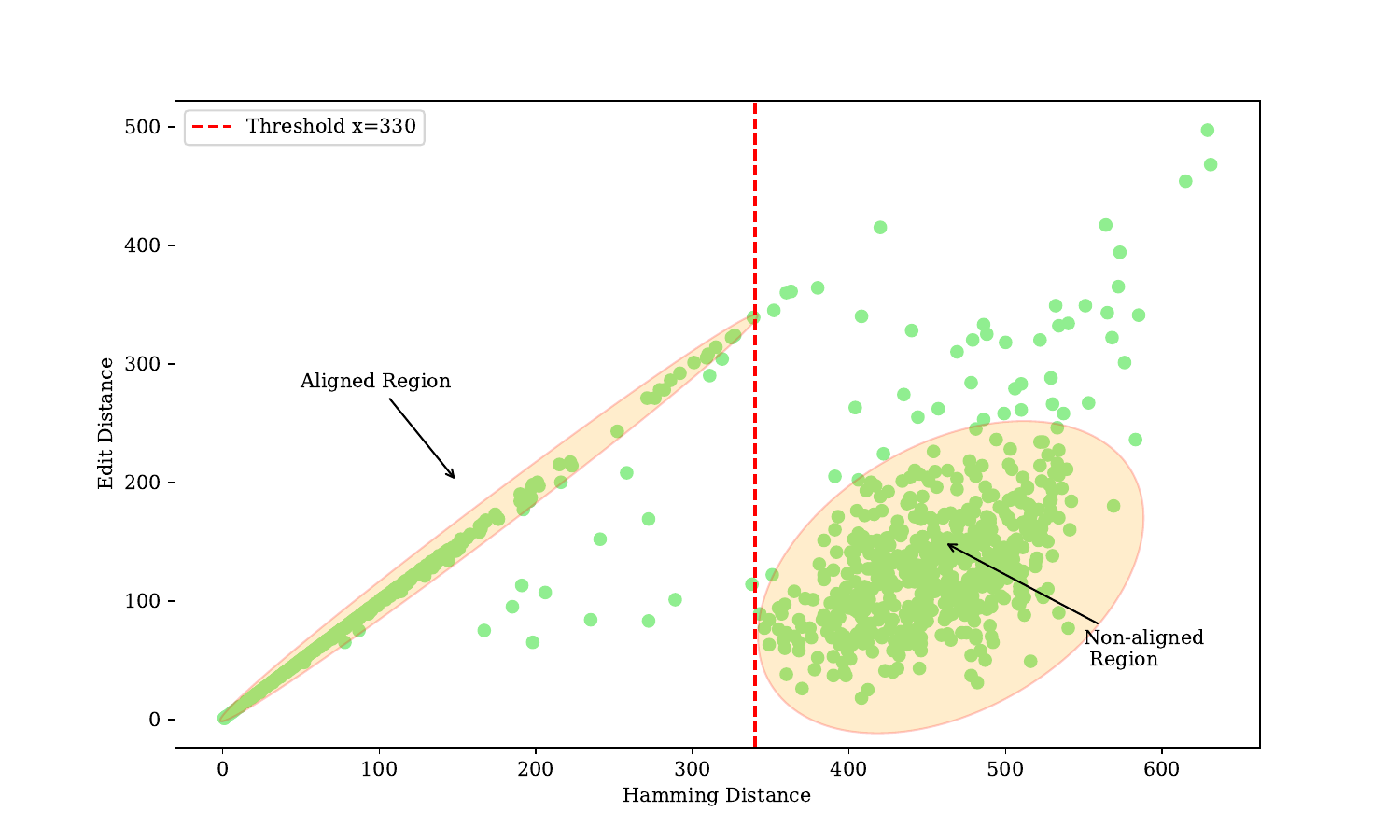}
\caption{Relationship between edit distance and Hamming distance for pairs of sequences sampled from different species across 1000 genera. The plot highlights two regions: the ``aligned region'', where the edit distance and Hamming distance are closely matched, and the ``non-aligned'' region, where misalignments result in greater Hamming distances relative to the edit distance.}
\label{fig:hamming_v_edit}
\end{figure}

To better understand the relationships between string distances and their practical utility, we considered 1000 different genera in the \textit{Pretraining} partition and sampled a pair of sequences from two different species within each genus. \autoref{fig:hamming_v_edit} illustrates how the edit distance varies with the Hamming distance in the experiment. Two salient regions are evident in the plot. In the ``aligned region'', the edit distance closely follows the Hamming distance as the number of mismatches is not due to alignment issues. In contrast, the ``non-aligned region'' potentially involves scenarios where sequences have been shifted, rearranged, or otherwise misaligned, leading to increased apparent mismatches when aligned positionally. These mismatches can be resolved with fewer string operations, resulting in a considerably lower edit distance compared to the Hamming distance.\\

The Hamming distance in the token space of an optimal tokenizer should closely follow the behaviour of the edit distance in both regions. This is an indicator of how much the tokenizer is helping in modelling the true sequence similarity. To study the behaviour of both tokenization strategies, we repeated the experiment with 1000 random genera from the pretraining dataset and visualized a scatter plot of the Hamming distance in token space versus the Hamming distance in sequence space for these sequences for both BPE and $k$-mer tokenizers.\\

\begin{figure}[bth]
\centering
\includegraphics[width=\linewidth]{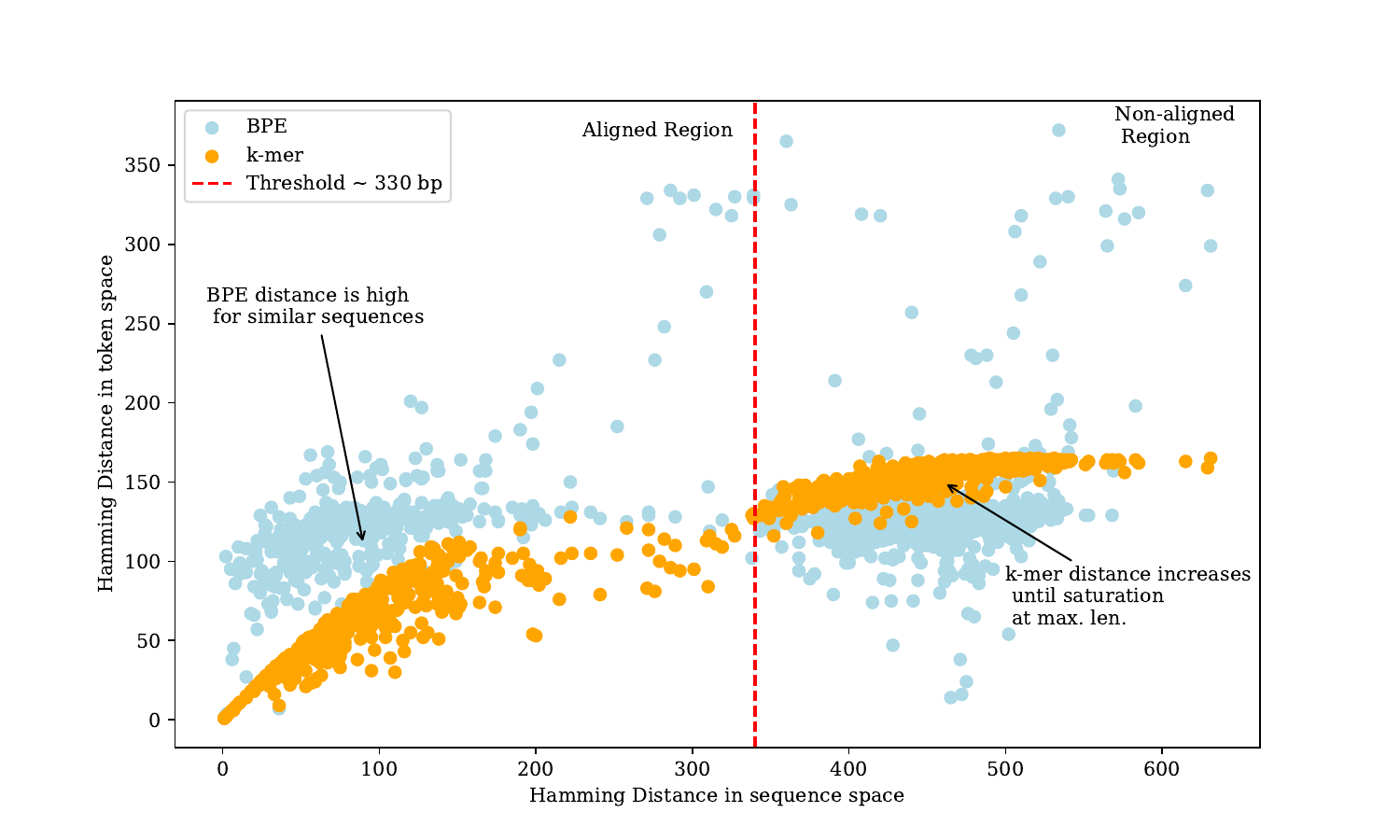}
\caption{Comparison of Hamming distances in token space versus sequence space for pairs of sequences sampled from different species across 1000 genera.  For both tokenizers, each dot in the plot represents the distances between sequences in a specific genus.}
\label{fig:tokenized_vs_hamming_dist}
\end{figure}

As illustrated in \autoref{fig:tokenized_vs_hamming_dist},  in the ``aligned region'', the $k$-mer tokenizer closely corresponds to the Hamming distance scaled by a factor of $1/k$. However, the BPE tokenizer is sensitive to minor changes in the sequences, possibly due to substitutions that lead to unknown tokens, causing the tokenizer to split these unknown tokens into single nucleotides. Even when the Hamming distance between two DNA sequences is small, their tokenized representations can differ significantly with BPE tokenization compared to $k$-mer tokenization.\\

On the other hand, the BPE tokenizer is more robust to frame shifts, following the behaviour of the edit distance in the ``non-aligned region'' of the plot. The $k$-mer tokenizer struggles in this region, as large Hamming distances can sometimes result from small frame shifts in DNA sequences, which is also illustrated in \autoref{fig:example_distances}. This limitation can be addressed by incorporating data augmentation during the pretraining phase to increase the model’s robustness to frame shifts by randomly offsetting the input sequence.\\

\autoref{tab:offset_ablations} shows the effect of pretraining with a random offset augmentation on the downstream genus-level accuracy for unseen species when using $k$-mer versus BPE tokenization with the BarcodeBERT model (4 layers, 4 attention heads). The $k$-mer length is set to $k\!=\!4$, and for both tokenizers the offset is randomly selected between 0 and 3 (inclusive). We use a weight of 1 for the substitution component of the loss function ($w_s\!=\!1$), a substitution token ratio of 50\% ($r_s \!=\! 50\%$), and the proportion of substitution tokens assigned to \texttt{[MASK]} set to 1 ($r_{\texttt{[MASK]}}\!=\!1$). As shown in \autoref{tab:offset_ablations}, random offset augmentation improves the genus-level accuracy of the $k$-mer-based model by 4.27 percentage points, but marginally \textit{decreases} the performance of the model trained with the BPE tokenizer.

\begin{table}[!htp]
\centering
\caption{Effect of augmenting with random offsets to the DNA sequences before tokenization with either $k$-mer or BPE tokenizers. In all experiments, we used a BarcodeBERT model with 4 layers and 4 attention heads, fixed weight for the substitution component of the loss function ($w_s\!=\!1$), substitution token ratio ($r_s\!=\! 50\%$) and substitution token proportion assigned to \texttt{[MASK]}, ($r_{\texttt{[MASK]}}\!=\!1$).}
\iftoggle{arxiv}{\small}{\footnotesize}
\begin{tabular}{lcc}
\toprule
% \multirow{2}{*}{Tokenizer} &\multirow{2}{*}{\makecell{ Random \\ Offset}} & \multirow{2}{*}{Linear probe } & \multirow{2}{*}{1-NN probe} & \multirow{2}{*}{ZSC-probe} \\
Tokenizer &\makecell{ Random \\ Offset} & 
\makecell{ Genus-level acc (\%) of \\ unseen species with $1$-NN probe} \\
% & & & & \\ 
\midrule
\multirow{2}{*}{k-mer}
 & \xmark     & 74.20\phantom{\tiny{{~${\mathord\uparrow}4.27$}}}\\
 & \checkmark & 78.47{\tiny{\textcolor{ForestGreen}{~$\mathord{\uparrow}4.27$}}} \\ \midrule
\multirow{2}{*}{BPE}
 & \xmark     & 69.85\phantom{\tiny{\textcolor{BrickRed}{~${\mathord\downarrow}1.29$}}} \\
 & \checkmark & 68.56{\tiny{\textcolor{BrickRed}{~$\mathord{\downarrow}1.29$}}} \\
\bottomrule
\end{tabular}
\label{tab:offset_ablations}
\end{table}

\clearpage
\section{Masking Strategy}
\label{a:masking}
In this supplementary experiment, we adapted our masking strategy to mimic BERT's original methodology \citep{Devlin2019BERTPO}. BERT addresses the difference in token distribution between pretraining and fine-tuning by employing a masking strategy whereby 80\% of substitution tokens are replaced by \texttt{[MASK]} tokens, 10\% are replaced with random tokens, and 10\% remain unchanged. This approach ensures more robust embeddings during testing, where masked tokens are absent \citep{Devlin2019BERTPO}. 

\begin{figure}[tbh]
    \centering
 \includegraphics[width=\iftoggle{arxiv}{0.75}{0.8}\linewidth]{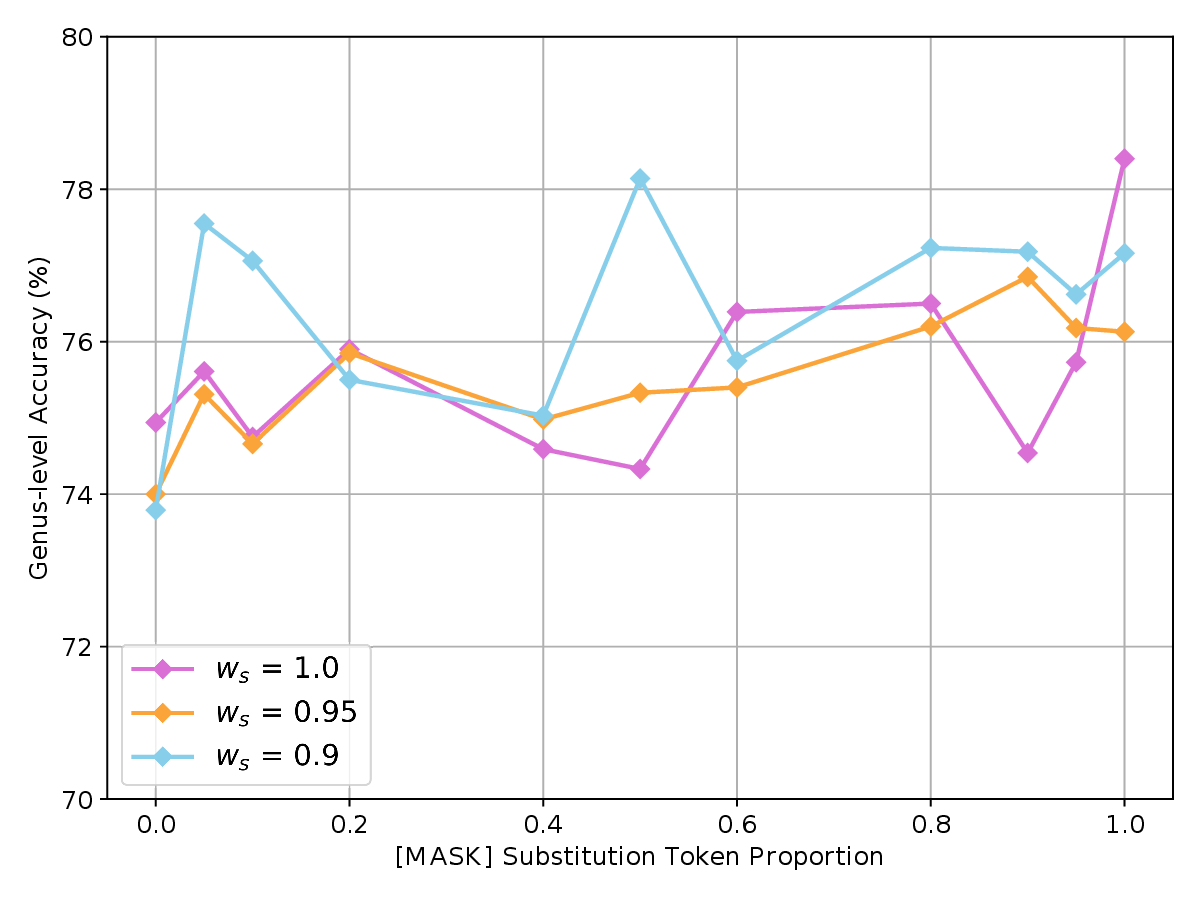}
    \caption{Genus-level accuracy for 1-NN probing of unseen species across different values of $r_{\texttt{[MASK]}}$, $r_{\texttt{[RNAD]}}$, and $w_s$. Experiments were conducted using the optimal configuration: 4 layers, 4 attention heads, $k\!=\!4$, with substitution token ratio ($r_s\!=\!50\%$).}
    \label{fig:mask_rand_proportion}
\end{figure}

To incorporate this methodology into BarcodeBERT, we define the following terms: $r_{\texttt{[MASK]}}$ as the proportion of substitution tokens assigned to the \texttt{[MASK]} token and $r_{\texttt{[RAND]}}\!=\!1-r_{\texttt{[MASK]}}$ as the proportion of the substitution tokens assigned a random valid token (all tokens except the special tokens). We explored various ratios for token replacement and three different values for the substitution loss penalty $w_s$. In the first case, based on the results of our experiments, where $w_s\!=\!1$ had the best performance, we kept $w_s\!=\!1$ and adjusted $r_{\texttt{[MASK]}}$. In the second case, we set $w_s$ to 0.95 and in the third case, $w_s$ was set to 0.90, to closely replicate BERT's original strategy that keeps 10\% of the tokens unchanged. Note that in all experiments $r_{\texttt{[RAND]}}$ was set to $1-r_{\texttt{[MASK]}}$.
In this study, we used the best configuration of 4 layers and 4 attention heads, $k\!=\!4$, and $r_s \!=\! 50\%$. \autoref{fig:mask_rand_proportion} presents the accuracy of these experiments for genus-level 1-NN probing on unseen species. The results show that for $w_s\!=\!1.0$, the best accuracy is 78.47\% with $r_{\texttt{[MASK]}} \!=\! 1.0$, for $w_s \!=\! 0.95$ the best accuracy is 76.85\% with $r_{\texttt{[MASK]}}\!=\!0.9$, and for $w_s \!=\! 0.9$, $r_{\texttt{[MASK]}}\!=\!0.5$ gives the best accuracy of 78.14\%, which improves the accuracy by 1\% compared to the case where $w_s \!=\! 0.9$ and $r_{\texttt{[MASK]}}\!=\!1.0$. Our results demonstrate that adopting BERT's masking strategy did not enhance the performance of BarcodeBERT, indicating that maintaining $r_{\texttt{[MASK]}} \!=\! 1$ is the optimal configuration.

\newpage
\section{Methodology --- Extra details}
\label{a:methodology}

To assess BarcodeBERT's performance, we employed a combination of alignment-based methods, transformer-based models, and clustering algorithms. Here, we describe the parameters and hardware used in each methodology for reproducibility. 

\subsection{BLAST}
We used BLAST \citep{altschul1990blast} for species-level classification to align query sequences from the test subset of the \textit{Seen} partition against a nucleotide database built from the training subset of the \textit{Seen} partition. To ensure high-quality alignments, we restricted the output to the top-scoring hit per query and applied strict thresholds of at least 80\% query coverage and 88\% sequence identity. For genus-level classification, we used BLAST to align query sequences from \textit{Unseen} partition against the same database built from the training subset of the \textit{Seen} partition.  We did not enforce a minimum coverage or sequence identity in this experiment as the expected sequence similarity at the genus level was lower than at the species level. We ran the experiments on an Intel(R) Xeon(R) CPU @ 2.20GHz using 4 CPU threads to maximize computational performance. The full commands used to run the experiments are as follows:
\vspace{0.2cm}
\begin{verbatim}
    blastn -query supervised_test.fas -db train.fas 
            -out results_supervised_top.tsv 
            -max_target_seqs 1 -qcov_hsp_perc 80 
            -perc_identity 88 -num_threads 4 
            -outfmt 6 

\end{verbatim}

\begin{verbatim}
    blastn -query unseen.fas -db train.fas 
            -out results_unseen.tsv -max_target_seqs 1 
            -num_threads 4 -outfmt 6 
\end{verbatim}

\subsection{Baseline Models}
\label{s:exp-dna-baseline}
For evaluation, we utilized the respective pretrained models from Hugging Face's ModelHub, specifically:

\begin{itemize}
\item DNABERT: \href{https://github.com/jerryji1993/DNABERT}{github.com/jerryji1993/DNABERT}
\item DNABERT-2: \href{https://huggingface.co/zhihan1996/DNABERT-2-117M}{huggingface.co/zhihan1996/DNABERT-2-117M}
\item DNABERT-S: \href{https://huggingface.co/zhihan1996/DNABERT-S}{huggingface.co/zhihan1996/DNABERT-S}
\item NT: \href{https://huggingface.co/InstaDeepAI/nucleotide-transformer-v2-50m-multi-species}{huggingface.co/InstaDeepAI/nucleotide-transformer-v2-50m-multi-species}
\item HyenaDNA: \href{https://huggingface.co/LongSafari/hyenadna-tiny-1k-seqlen-d256-hf}{huggingface.co/LongSafari/hyenadna-tiny-1k-seqlen-d256-hf}
\end{itemize}

\subsection{Model architecture configuration}

To identify the optimal model configuration, we conducted a series of ablation experiments evaluating the impact of architectural choices, tokenization strategies, and the substitution component of the loss ($w_s$). We considered $k$-mer tokenization with four distinct $k$ values (2, 4, 6, 8), as well as the BPE tokenizer. Three transformer configurations were evaluated: (i) 4 layers with 4 attention heads, (ii) 6 layers with 6 heads, and (iii) 12 layers with 12 heads. Additionally, we varied the penalty weight assigned to the substitution component of the loss ($w_s$) function, exploring values from 0.2 to 1.0.

\autoref{tab:loss-weight-experiment} reports genus-level accuracy for 1-NN probing of unseen species across these settings. The best performance was obtained with the 4-4-4 configuration (4-mers, 4 heads, 4 layers), which we adopt as the default model throughout our evaluations.

\begin{table*}[!htp]\centering
\scriptsize
\caption{Genus-level classification accuracy for 1-NN probing of unseen species using different $k$-mer sizes, transformer configurations, and substitution loss weights ($w_s$). Best results for each configuration are indicated in \first{boldface}.}
\label{tab:loss-weight-experiment}
\setlength{\tabcolsep}{3pt} % Adjust column padding
\renewcommand{\arraystretch}{1.1} % Adjust row padding
\begin{tabular*}{\textwidth}{@{\extracolsep{\fill}}crrrrrrrrrrrrrrr@{\extracolsep{\fill}}}
\toprule
& \multicolumn{15}{c}{Genus-level acc (\%)} \\
& \multicolumn{15}{c}{of unseen species with $1$-NN probe} \\
\cmidrule(l){2-16}
& \multicolumn{5}{c}{4 layers, 4 heads} & \multicolumn{5}{c}{6 layers, 6 heads} & \multicolumn{5}{c}{12 layers, 12 heads} \\
\cmidrule(l){2-6}\cmidrule(l){7-11}\cmidrule(l){12-16}
Loss weight ($w_s$) & $k\!=\!2$ &$k\!=\!4$ &$k\!=\!6$ &$k\!=\!8$ & BPE & $k\!=\!2$ &$k\!=\!4$ &$k\!=\!6$ &$k\!=\!8$ & BPE & $k\!=\!2$ &$k\!=\!4$ &$k\!=\!6$ &$k\!=\!8$ & BPE \\
\midrule

    0.2 & 64.18 & 76.06 & 75.15 & 71.15 & 70.57 & 61.59 & 74.61 & 70.87 & 67.74 & 67.15  & 48.92 & 63.72 & 57.12 & 56.40 & 62.34 \\
    0.5 & 66.47 & 74.98 & 76.62 & 71.22 & 70.34 & 65.38 & 73.37 & 70.87 & 69.70 &  67.57 & 46.23 & 67.11 & 61.24 & 60.05 & 62.20  \\
    0.8 & 68.84 & 76.71 & 74.66 & 73.33 & 69.40  &  68.37 & 74.02 & 72.20 & 69.72 & 68.23 & 60.50 & 67.50 &  66.87 & 61.05 & 67.09 \\
    1.0 & 76.92 & \first{78.47} & 75.74 & 75.62 & 69.85& 67.71 & 73.91 & 74.38 & \first{75.33} & 70.45 & \first{73.98} & 68.27 & 68.67 & 73.79 &  68.16 \\
\bottomrule
\end{tabular*}
\vspace{-0.5mm}
\end{table*}

\subsection{Pretraining}
BarcodeBERT was pretrained for 35 epochs using the AdamW optimizer \citep{AdamW} with a learning rate of $\alpha = 2\times10^{-4}$, a batch size of 128, and a OneCycle learning rate scheduler \citep{onecycle}. The pretraining process utilized four NVIDIA V100 GPUs and required approximately 36 hours to complete for each experiment executed. To examine the impact of pretraining, we also trained a model from scratch on the training subset of the \textit{Seen} partition without any pretraining. %While this model achieved high accuracy (99.3\%) on species-level classification, its performance on downstream tasks was suboptimal, highlighting the importance of pretraining for robust generalization.

\subsection{Fine-tuning}
All baseline models and BarcodeBERT were fine-tuned for 35 epochs on the supervised training subset of the \textit{Seen} partition. For larger models, a batch size of 32 was used, while smaller models (CNN, HyenaDNA and BarcodeBERT) were trained with a batch size of 128. The AdamW optimizer \citep{AdamW} with a learning rate of $\alpha = 1\times10^{-4}$ was employed, coupled with the OneCycle learning rate scheduler \citep{onecycle}. We used a single NVIDIA V100 GPU for the fine-tuning process, completed within 18 hours for most models. Notably, HyenaDNA required significantly less time to fine-tune compared to other models, due to its lightweight architecture. 

To train the fully-supervised BarcodeBERT model, which ablates the pretraining stage, we trained a randomly initialized model and then followed the same training procedure as our fine-tuning process.

\subsection{Linear probe training}
A linear classifier is applied to the embeddings generated by all the pretrained models for species-level classification. The models' parameters are learned using stochastic gradient descent with a constant learning rate of 1, momentum $\mu\!=\!0.95$, and weight decay $\lambda\!=\!1\times10^{-9}$.

\subsection{Zero-shot clustering}

We evaluated the models’ ability to group sequences without supervision using a modified version of the framework from \iftoggle{arxiv}{\citet{zsc-Lowe-2024}}{Lowe et. al. \cite{zsc-Lowe-2024}}. Embeddings were extracted from the pretrained encoders and reduced to 50 dimensions using UMAP \citep{umap} and cosine similarity to enhance computational efficiency while preserving data structure. These reduced embeddings were clustered with Agglomerative Clustering (L2 distance, Ward’s linkage), using the number of true species as the target number of clusters. Clustering performance was assessed with adjusted mutual information (AMI) to measure alignment with ground-truth labels.

\subsection{Additional classification metrics}
To complement our species‐ and genus‐level accuracy analyses, we also report the weighted F1‐score, which combines precision and recall together to create a single metric, computed (on our imbalanced barcode dataset) with sklearn's \texttt{average="weighted"} setting. The baselines include alignment‐based BLAST, a non‐SSL CNN encoder, and off‐the‐shelf DNA foundation models pretrained on generic genomic datasets. These are compared against BarcodeBERT (4-4-4), our model pretrained specifically on DNA barcode data.
% \best{Boldface} indicates the best performance for each metric, and \runnerup{underscores} denote the second best.
\begin{table*}[h!]
\caption{Weighted F1‐scores of different classification models under different evaluation strategies. 
Baselines are grouped into an alignment‐based method (BLAST), a non‐SSL CNN encoder, and generic DNA foundation models pretrained off‐the‐shelf. 
These are compared against BarcodeBERT (configured with $k\!=\!4$, 4 heads, 4 layers). 
{Boldface} indicates the \first{best result} per column, and {underlines} indicate \runnerup{second place}.}
\label{tab:f1_weighted}
\footnotesize
\centering
\begin{tabular}{lrrr}
\toprule
{Model} & \multicolumn{1}{r}{\makecell{{Species‐level F1 (\%)}\\{(seen species)}}} 
      & \multicolumn{1}{r}{\makecell{{Species‐level F1 (\%)}\\{(linear probe)}}} 
      & \multicolumn{1}{r}{\makecell{{Genus‐level F1 (\%)}\\{(1‐NN probe)}}} \\ 
\midrule
BLAST             &  99.66       & \na{}     & \first{81.77}     \\ 
\midrule
CNN encoder       &  98.54      &    \na{}       & 49.12    \\  % Unfair LP: 98.48
\midrule
DNABERT ($k\!=\!6$) & 99.53        &   \runnerup{98.48}       & 45.09        \\
DNABERT‐2         & \runnerup{99.70} &    95.61       & 21.59        \\
DNABERT‐S         & \first{99.74} &    96.85       & 48.00       \\
HyenaDNA‐tiny (d256)& 99.12        &    96.21       &   47.98      \\
Nucleotide Transformer  & 99.51  &    96.23       & 37.93        \\
\midrule
\textbf{BarcodeBERT (4-4-4)} & \first{99.74} & \first{99.34} & \runnerup{76.74} \\
\bottomrule
\end{tabular}
\end{table*}

\subsection{Additional performance metrics}
Inference-time resource consumption for each model is summarized in \autoref{tab:resoure_consumption}. CPU consumption is broadly consistent across all the foundation models, with HyenaDNA using around 3\% fewer resources than the other models. This state-space, attention-free model replaces quadratic self-attention with linear recurrence kernels, behaving like a CNN during inference and yielding minimal VRAM ($\sim$130\,MB) and host RAM (4.9\,GB) consumption. In contrast, models in the DNABERT family require over 11\,GB of system memory and 2.3--3.0\,GB of GPU memory to store large attention matrices. Both the Nucleotide Transformer and BarcodeBERT moderate those costs through different design choices. NT uses the evolutionary-scale modelling architecture (7.6\,GB RAM, 1.9\,GB VRAM), and BarcodeBERT uses a compact BERT encoder (6.5\,GB RAM, 0.7\,GB VRAM). BLAST remains the most efficient option, with zero VRAM usage and only 222\,MB of host RAM, reflecting its disk-backed, optimized indexing engine. Future work may include model compression (e.g. pruning, quantization) and optimized inference kernels to further narrow the efficiency gap among deep learning approaches.

\begin{table}[ht]
\centering
\footnotesize
\begin{tabular}{lrrr}
\toprule
Model & {CPU utilization (\%)} & {RAM (MB)} & {VRAM (MB)} \\
\midrule
BLAST                    & 7.26  &  222 & 0 \\
\midrule
CNN encoder              &  22.11 & 5,332 & 98 \\
\midrule
DNABERT ($k\!=\!6$)      &  24.91 & 11,345  & 2,285 \\
DNABERT-2                &  25.01 & 12,669 & 3,010 \\
DNABERT-S                & 25.01  & 12,696 & 3,009 \\
HyenaDNA-tiny (d256)            & 22.18  & 4,851 & 131 \\
Nucleotide Transformer   & 24.96  & 7,607  & 1,885 \\
\midrule
BarcodeBERT   & 25.42  & 6,493 &  694 \\
\bottomrule
\end{tabular}
\caption{Resource usage comparison of DNA models. CPU usage is normalized to a 4-core system.}
\label{tab:resoure_consumption}
\end{table}

\end{document}